\documentclass[runningheads]{llncs}

\usepackage[mobile]{eccv}

\usepackage{color}
\usepackage{colortbl}
\usepackage{multirow}
\usepackage{mathtools}
\usepackage{cancel}
\newcommand\ExtraSep
{\dimexpr\cmidrulewidth+\aboverulesep+\belowrulesep\relax}
\usepackage{eccvabbrv}

\usepackage{graphicx}
\usepackage{booktabs}

\usepackage[accsupp]{axessibility}  

\usepackage{amsmath}
\usepackage[export]{adjustbox}

\usepackage[pagebackref,breaklinks,colorlinks,citecolor=eccvblue]{hyperref}
\usepackage{hyperref}

\usepackage{orcidlink}

\newcommand{\bB}{\mathbf{B}}
\newcommand{\bc}{\mathbf{c}}\newcommand{\bC}{\mathbf{C}}

\newcommand{\bG}{\mathbf{G}}

\newcommand{\bI}{\mathbf{I}}

\newcommand{\bo}{\mathbf{o}}

\newcommand{\bq}{\mathbf{q}}
\newcommand{\bR}{\mathbf{R}}
\newcommand{\bS}{\mathbf{S}}
\newcommand{\bt}{\mathbf{t}}\newcommand{\bT}{\mathbf{T}}
\newcommand{\bu}{\mathbf{u}}

\newcommand{\bx}{\mathbf{x}}

\newcommand{\bvarepsilon}{\boldsymbol{\varepsilon}}

\newcommand{\btheta}{\boldsymbol{\theta}}

\newcommand{\bmu}{\boldsymbol{\mu}}

\newcommand{\bSigma}{\boldsymbol{\Sigma}}

\newcommand{\nR}{\mathbb{R}}

\newcommand{\cL}{\mathcal{L}}

\newcommand{\figref}[1]{Fig.~\ref{#1}}

\newcommand{\eqnref}[1]{Eq.~\eqref{#1}}

\makeatletter
\DeclareRobustCommand\onedot{\futurelet\@let@token\@onedot}
\def\@onedot{\ifx\@let@token.\else.\null\fi\xspace}
 
\def\ie{i.e\onedot}

\makeatother

\newcommand{\PAR}[1]{\vspace{0.1cm}\noindent{\bf #1} }

\begin{document}

\title{BAD-Gaussians: Bundle Adjusted Deblur Gaussian Splatting} 


\author{Lingzhe Zhao\inst{1*}\orcidlink{0009-0005-8000-1525} \and
Peng Wang\inst{1,2*}\orcidlink{0009-0003-5747-3319} \and
Peidong Liu\inst{1}\orcidlink{0000-0002-9767-6220}}

\authorrunning{Lingzhe Zhao, Peng Wang, Peidong Liu}

\institute{$^{1}$Westlake University \quad $^{2}$Zhejiang University \\
\email{\{zhaolingzhe,wangpeng,liupeidong\}@westlake.edu.cn}\\
\url{https://lingzhezhao.github.io/BAD-Gaussians/}}

\maketitle

\renewcommand{\thefootnote}{\fnsymbol{footnote}}
\footnotetext[1]{Equal contribution.}

\begin{abstract}
While neural rendering has demonstrated impressive capabilities in 3D scene reconstruction and novel view synthesis, it heavily relies on high-quality sharp images and accurate camera poses. Numerous approaches have been proposed to train Neural Radiance Fields (NeRF) with motion-blurred images, commonly encountered in real-world scenarios such as low-light or long-exposure conditions. However, the implicit representation of NeRF struggles to accurately recover intricate details from severely motion-blurred images and cannot achieve real-time rendering. In contrast, recent advancements in 3D Gaussian Splatting achieve high-quality 3D scene reconstruction and real-time rendering by explicitly optimizing point clouds into 3D Gaussians.
%
In this paper, we introduce a novel approach, named BAD-Gaussians (Bundle Adjusted Deblur Gaussian Splatting), which leverages explicit Gaussian representation and handles severe motion-blurred images with inaccurate camera poses to achieve high-quality scene reconstruction. Our method models the physical image formation process of motion-blurred images and jointly learns the parameters of Gaussians while recovering camera motion trajectories during exposure time.
%
In our experiments, we demonstrate that BAD-Gaussians not only achieves superior rendering quality compared to previous state-of-the-art deblur neural rendering methods on both synthetic and real datasets but also enables real-time rendering capabilities.
\end{abstract}
\section{Introduction}
\label{sec:intro}
%
Acquiring accurate 3D scene representations from 2D images has long been a challenging problem in computer vision. Serving as a fundamental component in various applications such as virtual/augmented reality and robotics navigation, substantial efforts have been dedicated to addressing this challenge over the last few decades. Among those pioneering works, Neural Radiance Fields (NeRF) \cite{nerf}, parameterized by Multi-layer Perceptrons (MLP), stands out for its utilization of differentiable volume rendering technique \cite{levoy1990efficient, max1995optical} and has garnered significant attention due to its capability to recover high-quality 3D scene representation from 2D images.
Numerous works have focused on enhancing the performance of NeRF, particularly in terms of training \cite{mueller2022instant, Chen2022ECCV, yu_and_fridovichkeil2021plenoxels} and rendering efficiency \cite{Garbin2021, Yu2021}. A recent advancement, 3D Gaussian Splatting (3D-GS) \cite{kerbl3Dgaussians}, extends the implicit neural rendering \cite{nerf} to explicit point clouds. By projecting these optimized point clouds (Gaussians) onto the image plane, 3D-GS \cite{kerbl3Dgaussians} achieves real-time rendering while enhancing the efficiency of NeRF in both training and rendering, and also improves rendering quality.
%
However, both NeRF-based \cite{nerf, mipnerf, mueller2022instant} methods and 3D-GS \cite{kerbl3Dgaussians} heavily rely on well-captured sharp images and accurately pre-computed camera poses, typically obtained from COLMAP \cite{colmap}. Motion-blurred images, a common form of image degradation, often encountered in low-light or long-exposure conditions, can notably impair the performance of both NeRF and 3D-GS. The challenges posed by motion-blurred images to NeRF and 3D-GS can be attributed to three primary factors: (a) NeRF and 3D-GS rely on high-quality sharp images for supervision. However, motion-blurred images violate this assumption and exhibit notably inaccurate corresponding geometry between multi-view frames, thus presenting significant difficulties in achieving accurate 3D scene representation for both NeRF and 3D-GS; (b) Accurate camera poses are essential for training NeRF and 3D-GS. However, recovering accurate poses from multi-view motion-blurred images using COLMAP \cite{colmap} is challenging. (c) 3D-GS necessitates sparse cloud points from COLMAP as the initialization of Gaussians. The mismatched features between multi-view blurred images and the inaccuracies in pose calibration further exacerbate the issue, leading to COLMAP producing fewer cloud points. This introduces an additional initialization issue for 3D-GS. Therefore, these factors result in a notable drop in performance for 3D-GS when dealing with motion-blurred images.
%

Implicit neural representations, \ie NeRF \cite{nerf}, have been employed to reconstruct sharp 3D scenes from motion-blurred images \cite{deblur-nerf, Lee_2023_CVPR, wang2023badnerf}. For example, Deblur-NeRF \cite{deblur-nerf} introduces a deformable sparse kernel that alters a canonical kernel at spatial locations to simulate the blurring process. DP-NeRF \cite{Lee_2023_CVPR} integrates physical priors derived from the motion-blurred image acquisition process into Deblur-NeRF \cite{deblur-nerf} to construct a clean NeRF representation. In contrast, BAD-NeRF \cite{wang2023badnerf} models the physical process of capturing motion-blurred images and jointly optimizes NeRF while recovering the camera trajectory within the exposure time. However, these implicit deblur rendering methods encounter significant challenges in achieving real-time rendering and producing high-quality outputs with intricate details. Additionally, the implicit representation introduces extra difficulties in optimizing the neural parameters and camera poses as mentioned in \cite{fu2023colmap}.
%

In order to address these challenges, we propose {\textbf{B}}undle {\textbf{A}}djusted {\textbf{D}}eblur {\textbf{Gaussian Splatting}}, the first motion deblur framework based on 3D-GS, which we refer to as {\textbf{BAD-Gaussians}}. We incorporate the physical process of motion blur into the training of 3D-GS, employing a spline function to characterize the trajectory within the camera's exposure time.  In the training of BAD-Gaussians, the camera trajectory within exposure time is optimized using gradients derived from the Gaussians of the scene, while jointly optimizing the Gaussians themselves.
%
Specifically, the trajectory of each motion-blurred image is represented by the initial and final poses at the beginning and end of the exposure time, respectively.  By assuming the exposure time is typically short, we can interpolate between the initial and final poses to obtain every camera pose along the trajectory. From this trajectory, we generate a sequence of virtual sharp images by projecting the scene's Gaussians onto the image plane. These virtual sharp images are then averaged to synthesize the blurred images, following the physical blur process. Finally, the Gaussians along the trajectory are optimized by minimizing the photometric error between the synthesized blurred images and the input blurred images through differentiable Gaussian rasterization.

We evaluate BAD-Gaussians using both synthetic and real datasets. The experimental results demonstrate that BAD-Gaussians outperforms prior state-of-the-art implicit neural rendering methods by explicitly incorporating the image formation process of motion-blurred images into the training of 3D-GS, achieving better rendering performance in terms of real-time rendering speed and superior rendering quality.
%
In summary, our {\bf{contributions}} can be outlined as follows:
\begin{itemize}
	\item We introduce a photometric bundle adjustment formulation specifically designed for motion-blurred images, achieving the first real-time rendering performance from motion-blurred images within the framework of 3D Gaussian Splatting;
	\item We demonstrate how this formulation enables the acquisition of high-quality 3D scene representation from a set of motion-blurred images;
	\item Our approach successfully deblurs severe motion-blurred images, synthesizes higher-quality novel view images, and achieves real-time rendering, surpassing previous state-of-the-art implicit deblurring rendering methods.
\end{itemize}

\section{Related Work}
\label{sec:related}
\subsection{Neural Radiance Fields}
NeRF \cite{nerf}, employing implicit MLP, exhibits remarkable performance in synthesizing high-quality novel view images and accurately representing 3D scenes. Numerous extension works have been proposed to enhance NeRF's performance, including improvements in training \cite{mueller2022instant, Chen2022ECCV, yu_and_fridovichkeil2021plenoxels, SunSC22} and rendering \cite{Garbin2021, Yu2021, Lassner2021, Piala2022, Wizadwongsa2021} efficiency, as well as anti-alias capabilities \cite{mipnerf, barron2022mipnerf360, barron2023zipnerf}. Additionally, various methods aim to bolster NeRF's robustness against imperfect inputs, such as inaccurate camera poses \cite{Lin2021, wang2021nerfmm, park2023camp, Jeong2021}, few-shot images \cite{infonerf, regnerf, depthnerf}, and low-quality images \cite{mildenhall2022rawnerf, deblur-nerf, Lee_2023_CVPR, wang2023badnerf, li2023usb}.
In the following section, we will primarily focus on reviewing methods closely related to our work.
\PAR{Fast Neural Rendering.}
Numerous approaches inspired by NeRF \cite{nerf} have sought to enhance its rendering efficiency by employing advanced data structures to reconstruct radiance fields, thereby minimizing the computational cost associated with implicit MLPs used in NeRF. These approaches are primarily categorized into grid-based \cite{Chen2022ECCV, yu_and_fridovichkeil2021plenoxels, Cao2023HEXPLANE, kplanes_2023} and hash-based \cite{mueller2022instant} methods. Despite these efforts, achieving real-time rendering for unbounded and complete scenes, as well as high-resolution images, remains challenging. In contrast to methods based on volume rendering and implicit representations, which may hinder fast rendering, recent advancements like 3D-GS \cite{kerbl3Dgaussians} achieve real-time high-quality rendering through pure explicit point scene representation and the differentiable Gaussian rasterization. Nevertheless, these 3D scene representations heavily rely on accurately posed high-quality images.
\PAR{NeRF for Camera Optimization.}
The pioneering work, BARF \cite{Lin2021}, was the first to propose simultaneous optimization of camera parameters alongside NeRF. They employed a coarse-to-fine bundle adjustment strategy to enhance camera pose recovery. Concurrently, SC-NeRF \cite{Jeong2021} introduced a method to learn various camera models, encompassing both extrinsic and intrinsic parameters, along with scene representation. In the latest development, CamP \cite{park2023camp} introduced a preconditioner to mitigate correlations between camera parameters, aiming for improved joint optimization of camera and NeRF parameters. In contrast to the aforementioned works \cite{Lin2021, Jeong2021, park2023camp}, which focus solely on optimizing camera poses for sharp images, our approach goes further by recovering the trajectory of each blurred image within the exposure time.
\PAR{NeRF for Deblurring.}
Several scene deblurring methods based on NeRF have been proposed, including Deblur-NeRF \cite{deblur-nerf} and DP-NeRF \cite{Lee_2023_CVPR}, which reconstruct sharp scene representations from sets of motion-blurred images. However, these methods fix inaccurate camera poses recovered from blurred images during training, leading to degradation in reconstruction performance. BAD-NeRF \cite{wang2023badnerf} jointly learns camera motion trajectories within exposure time and radiance fields, following the physical blur process. Despite these advancements, existing deblurring neural rendering strategies struggle to achieve real-time rendering and reconstruct intricate scene details due to the implicit MLP structure, necessitating significant improvements in rendering efficiency and quality. Moreover, optimizing the 3D scenes jointly with camera poses faces additional challenges due to NeRF's implicit representations. To address these issues, we propose achieving deblurring capability within the framework of Gaussian Splatting \cite{kerbl3Dgaussians}.

\vspace{-1em}
\subsection{Image Deblurring}
Two primary categories typically classify existing techniques for addressing the motion deblurring problem: the first involves formulating the issue as an optimization task, wherein gradient descent is employed during inference to jointly refine the blur kernel and the latent sharp image \cite{Cho2009ToG,Fergus2006SIGGGRAPH,Krishnan2009NIPS,Levin2009CVPR,Shan2008ToG,Xu2010ECCV,Park2017}. Another pipeline phrases the deblurring task as an end-to-end learning, particularly leveraging deep learning methodologies. With the support of substantial datasets and deep neural networks, superior results have been achieved for both single image deblurring \cite{Nah2017CVPR,Tao2018CVPR,Kupyn2019ICCV} and video deblurring \cite{Su2017CVPR}.
However, these 2D deblurring methods cannot exploit the 3D scene geometry between multi-view images, thus failing to ensure the view consistency of the scene from different viewpoints. In contrast, our approach focuses on leveraging the geometry within multi-view blurry images to reconstruct a high-quality 3D scene.

\section{Method}
\label{sec:method}

\begin{figure*}[!t]
	\centering
	\includegraphics[width=1.0\linewidth]{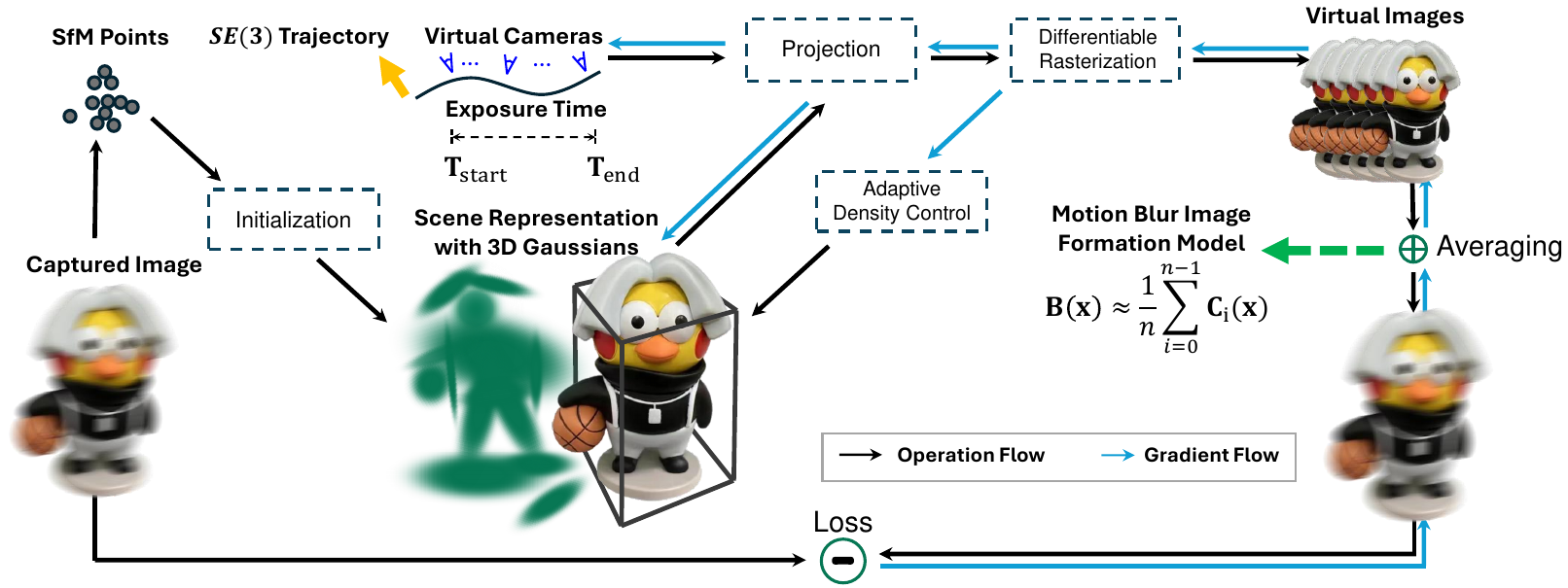}
	\caption{{\bf{The pipeline of BAD-Gaussians.}} Our approach utilizes Gaussian representations to depict sharp 3D scenes derived from a series of motion-blurred images, along with their inaccurate poses and sparse point clouds from COLMAP, serving as the initialization for the Gaussians. Employing forward projection and differentiable Gaussian rasterization, we jointly optimize the Gaussians in the scene and the camera trajectory within exposure time, by backpropagating gradients from Gaussians to camera poses. Following the physical process of motion blur, we model motion-blurred images by averaging the virtual sharp images captured during the exposure time. These virtual camera poses are represented and interpolated using a continuous spline within the $\mathbf{SE}(3)$ space. The joint optimization of Gaussians and camera trajectories is achieved by minimizing the photometric loss between synthesized and actual blurry images.}
	\label{fig_overview}
	\vspace{-1.5em}
\end{figure*}

BAD-Gaussians aims to recover a sharp 3D scene representation by jointly learning the camera motion trajectories and Gaussians parameters, given a sequence of motion-blurred images, along with their inaccurate poses and sparse point clouds estimated from COLMAP \cite{colmap}, as shown in Fig. \ref{fig_overview}. This is achieved by minimizing the photometric error between the input blurred images and the synthesized blurred images generated based on the physical motion blur image formation model. We will deliver each content in the following sections.

\subsection{Preliminary: 3D Gaussian Splatting}
\label{3d-gs}
Following 3D-GS \cite{kerbl3Dgaussians}, the scene is represented by a series of 3D Gaussians. Each Gaussian, denoted as $\bG$, is parameterized by its mean position $\boldsymbol{\mu} \in \mathbb{R}^3$, 3D covariance $\mathbf{\Sigma} \in \mathbb{R}^{3 \times 3}$, opacity $\bo \in \mathbb{R} $ and color $\bc \in \mathbb{R}^3$. The distribution of each scaled Gaussian is defined as:
\begin{equation}
	\bG(\bx) = e^{-\frac{1}{2}(\bx-\bmu)^{\top}\mathbf{\Sigma}^{-1}(\bx-\bmu)}.
	\label{eq:gauss}
\end{equation}

To ensure that the 3D covariance $\mathbf{\Sigma}$ remains positive semi-definite, which is physically meaningful, and to reduce the optimization difficulty, 3D-GS represents $\mathbf{\Sigma}$ using a scale $\bS \in \mathbb{R}^3$ and rotation matrix $\bR \in \mathbb{R}^{3 \times 3}$ stored by a quaternion $\bq \in \mathbb{R}^4$:
\begin{equation}
	\mathbf{\Sigma} = \mathbf{RSS}^{T}\mathbf{R}^{T}.
	\label{eq:3dcovariance}
\end{equation}

In order to enable differentiable Gaussian rasterization, 3D-GS projects 3D Gaussians to 2D from a given camera pose $\mathbf{T}_c = \{\mathbf{R}_c \in \mathbb{R}^{3 \times 3}, \mathbf{t}_c \in \mathbb{R}^3\}$ for rasterizing and rendering using the following equation, as described in \cite{zwicker2001ewa}:
\begin{equation}
	\mathbf{\Sigma^{\prime}} = \mathbf{JR}_c\mathbf{\Sigma R}_c^T\mathbf{J}^{T},
	\label{eq:covariance2d}
\end{equation}
where $\mathbf{\Sigma^{\prime}} \in \mathbb{R}^{2 \times 2}$ is the 2D covariance matrix, $\mathbf{J} \in \mathbb{R}^{2 \times 3}$ is the Jacobian of the affine approximation of the projective transformation.
Afterward, each pixel color is rendered by rasterizing these $N$ sorted 2D Gaussians based on their depths, following the formulation:
\begin{equation}
	\bC = \sum_{i}^{N} \bc_i \alpha_i \prod_j^{i-1}(1-\alpha_j),
	\label{eq:render}
\end{equation}
where $\bc_i$ is the learnable color of each Gaussian, and $\alpha_i$ is the alpha value computed by evaluating a 2D covariance $\mathbf{\Sigma^{\prime}}$ multiplied with the learned Gaussian opacity $\bo$:
\begin{align}
	\alpha_i = \bo_i \cdot \exp(-\sigma_i), \quad
	\sigma_i = \frac{1}{2} {\rm \Delta}_i^T \mathbf{\Sigma^{\prime}}^{-1} {\rm \Delta}_i,
	\label{eq_alpha}
\end{align}
where ${\rm \Delta} \in \mathbb{R}^2$ is the offset between the pixel center and the 2D Gaussian center.
%

The derivations presented above demonstrate that the rendered pixel color, denoted as $\bC$ in \eqnref{eq:render}, is a function that is differentiable with respect to all of the learnable Gaussians $\bG$, and the camera poses $\bT_c$. This facilitates our bundle adjustment formulation, accommodating a set of motion-blurred images and inaccurate camera poses within the framework of 3D-GS.

\subsection{Physical Motion Blur Image Formation Model}
\label{blurmodel}

The physical process of image formation in a digital camera encompasses the gathering of photons during the exposure period, followed by their conversion into measurable electric charges. Mathematically representing this phenomenon involves integrating across a sequence of simulated virtual latent sharp images, as follows:
\begin{equation}\label{eq_continuous_blur_im_formation}
	\bB(\bu) = \phi \int_{0}^{\tau} \bC_\mathrm{t}(\bu) \mathrm{dt}, \quad
\end{equation}
where $\bB(\bu) \in \nR^{\mathrm{H} \times \mathrm{W} \times 3}$ denotes the real captured motion-blurred image, $\bu \in \nR^2$ represents the pixel location in the image with height $\mathrm{H}$ and width $\mathrm{W}$, $\phi$ serves as a normalization factor, $\tau$ is the camera exposure time, $\bC_\mathrm{t}(\bu) \in \nR^{\mathrm{H} \times \mathrm{W} \times 3}$ is the virtual latent sharp image captured at timestamp $\mathrm{t} \in [0, \tau]$ within the exposure time. The blurred image $\bB(\bu)$, resulting from camera motion during the exposure time, is calculated by averaging all the virtual images $\bC_\mathrm{t}(\bu)$ across different timestamps $t$.  This discrete approximated of this model is dependent on the number $n$ of discrete samples, as denoted: 
\begin{equation}\label{eq_blur_im_formation}
	\bB(\bu)  \approx \frac{1}{n} \sum_{i=0}^{n-1} \bC_i(\bu). 
\end{equation}

The level of motion blur within an image is contingent upon the movement of the camera during the exposure time. For instance, a swiftly moving camera results in minimal relative motion, particularly with shorter exposure times, while a slowly moving camera yields motion-blurred images, especially in low-light scenarios with prolonged exposure times. Additionally, it can be deduced that $\bB(\bu)$ demonstrates differentiability concerning each virtual sharp image $\bC_i(\bu)$.

\subsection{Camera Motion Trajectory Modeling in 3D-GS}
\label{trajectory_in_3dgs}
Based on \eqnref{eq_blur_im_formation}, a straightforward approach to address motion-blurred images involves determining each virtual sharp image $\bC_i$, which serves as the dependent variable of the motion-blur image $\bB$. Given that a sharp image $\bC_i$ can be rendered from a specified camera pose $\bT_i$ within the framework of 3D-GS (\ie $\mathbf{G}_\theta$) \cite{kerbl3Dgaussians}, establishing a one-to-one correspondence between poses and virtual sharp images is feasible. Consequently, we formulate the corresponding poses of each latent sharp image within the exposure time $\tau$ by employing a camera motion trajectory represented through linear interpolation between two camera poses, one at the beginning of the exposure $\bT_\mathrm{start} \in \mathbf{SE}(3)$ and the other at the end $\bT_\mathrm{end} \in \mathbf{SE}(3)$. The virtual camera pose at time $t \in [0,\tau]$ can thus be expressed as follows:
\begin{equation} \label{eq_trajectory}
	\bT_t = \bT_\mathrm{start} \cdot \mathrm{exp}(\frac{t}{\tau} \cdot \mathrm{log}(\bT_\mathrm{start}^{-1} \cdot \bT_\mathrm{end})),
\end{equation} 
where $\tau$ represents the exposure time. $\frac{t}{\tau}$ can be further discretized and derived as $\frac{i}{n-1}$ for the $i^{th}$ sampled virtual sharp image (\ie $C_i$) in \eqnref{eq_blur_im_formation} (i.e. with pose denoted as $\bT_i$), when there are $n$ sampled images in total. It follows that $\bT_i$ is differentiable with respect to both $\bT_{\mathrm{start}}$ and $\bT_{\mathrm{end}}$. We refer to prior works \cite{mba-vo, wang2023badnerf} for a comprehensive understanding of the interpolation and derivations of the related Jacobian. \cite{mba-vo}. The objective of BAD-Gaussians is to estimate both $\bT_\mathrm{start}$ and $\bT_\mathrm{end}$ for each frame, along with the learnable parameters of Gaussians $\mathbf{G}_\theta$. 


\subsection{Loss Function}
\label{loss_function}
From a collection of $K$ motion-blurred images, we can proceed to estimate both the learnable parameters $\btheta$ (\ie mean position $\boldsymbol{\mu}$, 3D covariance $\mathbf{\Sigma}$, opacity $\bo$ and color $\bc$) of 3D-GS and the camera motion trajectories (\ie $\bT_\mathrm{start}$ and $\bT_\mathrm{end}$) for each image. This estimation is accomplished by minimizing the following loss function, which includes an $\mathcal{L}_1$ loss and a D-SSIM term between $\bB_k(\bu)$, the $k^{th}$ blurry image synthesized from 3D-GS using the aforementioned image formation model (\ie \eqnref{eq_blur_im_formation}), and $\bB^{gt}_k(\bu)$, the corresponding real captured blurry image:
\begin{equation}
	\cL = (1-\lambda) \mathcal{L}_1 + \lambda \mathcal{L}_{\text{D-SSIM}}.
\end{equation} 

To optimize the learnable Gaussians parameter $\btheta$, and camera pose $\bT$ (\ie $\bT_\mathrm{start}$ and $\bT_\mathrm{end}$ in our model) for each image, it is necessary to derive the corresponding Jacobians to complete the gradient flow:
\begin{equation}
	\frac{\partial \cL}{\partial \btheta} = \sum_{k=0}^{K-1} \frac{\partial \cL}{\partial \bB_k} 
	\cdot \frac{1}{n}\sum_{i=0}^{n-1} \frac{\partial \bB_k}{\partial \bC_i} 
	\frac{\partial \bC_i}{\partial \btheta},
    \label{eq:grad_theta}
\end{equation}
\begin{equation}
	\frac{\partial \cL}{\partial \bT} = \sum_{k=0}^{K-1} \frac{\partial \cL}{\partial \bB_k} 
	\cdot \frac{1}{n}\sum_{i=0}^{n-1} \frac{\partial \bB_k}{\partial \bC_i} \frac{\partial \bC_i}{\partial \btheta}
	\frac{\partial \btheta}{\partial \bT},
    \label{eq:grad_pose}
\end{equation}
%
where $\bB_k(\bu)$ and $\bC_i(\bu)$ are denoted as $\bB_k$ and $\bC_i$ for simplification, and we parameterize both $\bT_\mathrm{start}$ and $\bT_\mathrm{end}$ with their corresponding Lie algebras of $\mathbf{SE}(3)$, which can be represented by a 6D vector respectively. For further details on the Jacobian of Gaussians to camera pose, $\frac{\partial \btheta}{\partial \bT}$, please refer to our supplemental materials.

\section{Experiments}
\label{sec:experiment}

\begin{table}[t]
	\centering
	\caption{{\bf{Ablation studies on the number of virtual camera poses $n$.}} The results indicate that performance reaches a saturation point with increasing number $n$.}
	\label{tab:ablation_n}
	\setlength\tabcolsep{6pt}{
		\resizebox{0.65\linewidth}{!}{
		\begin{tabular}{c|ccc|ccc}
			\hline
			& \multicolumn{3}{c|}{Cozy2room} & \multicolumn{3}{c}{Tanabata}\\
			
			$n$ & {\scriptsize PSNR$\uparrow$} & {\scriptsize SSIM$\uparrow$} & {\scriptsize LPIPS$\downarrow$} & {\scriptsize PSNR$\uparrow$} & {\scriptsize SSIM$\uparrow$} & {\scriptsize LPIPS$\downarrow$} \\
			\specialrule{0.05em}{1pt}{1pt}
			3 & 31.63 & .9389 & .0810 & 25.43 & .8215 & .2287 \\
			
			4 & 32.99 & .9552 & .0558 & 27.08 & .8704 & .1764 \\
			
			5 & 33.68 & .9595 & .0458 & 28.39 & .9011 & .1400 \\
			
			6 & 33.86 & .9627 & .0368 & 29.57 & .9214 & .1101 \\
			
			8 & 34.48 & \cellcolor{orange!25}.9654 & .0315 & 31.15 & .9440 & .0696 \\
			
			10 & \cellcolor{orange!25}34.68 & .9521 & \cellcolor{orange!25}.0258 & \cellcolor{yellow!25}32.12 & \cellcolor{yellow!25}.9481 & \cellcolor{yellow!25}.0464 \\
			
			15 & \cellcolor{red!25}34.78 & \cellcolor{red!25}.9679 & \cellcolor{red!25}.0257 & \cellcolor{orange!25}33.58 & \cellcolor{orange!25}.9652 & \cellcolor{orange!25}.0198 \\
			
			20 & \cellcolor{yellow!25}34.63 & \cellcolor{yellow!25}.9646 & \cellcolor{yellow!25}.0307 & \cellcolor{red!25}33.87 & \cellcolor{red!25}.9668 & \cellcolor{red!25}.0151 \\
			\hline
		\end{tabular}
	}
	}

	\vspace{1.2em}

	\setlength\tabcolsep{2pt}{
	\caption{{\bf{Ablation studies on the effect of trajectory representations.}} We denote Deblur-NeRF-S and Deblur-NeRF-R as the synthetic and real data from Deblur-NeRF, respectively. The results demonstrate that cubic interpolation improves performance in scenes with complex camera trajectories (\ie {\it{MBA-VO}} and {\it{Deblur-NeRF-R}}).}
	\resizebox{0.95\linewidth}{!}{
		\begin{tabular}{c|ccc|ccc|ccc}
			\hline
			& \multicolumn{3}{c|}{\it{Deblur-NeRF-S}}& \multicolumn{3}{c|}{\it{MBA-VO}} &  \multicolumn{3}{c}{\it{Deblur-NeRF-R}} \\
			& {\scriptsize PSNR$\uparrow$} & {\scriptsize SSIM$\uparrow$} & {\scriptsize LPIPS$\downarrow$} & {\scriptsize PSNR$\uparrow$} & {\scriptsize SSIM$\uparrow$} & {\scriptsize LPIPS$\downarrow$} & {\scriptsize PSNR$\uparrow$} & {\scriptsize SSIM$\uparrow$} & {\scriptsize LPIPS$\downarrow$} \\
			\hline
			Linear Interpolation& \textbf{33.92} & \textbf{.9467} & \textbf{.0422} & 30.96 & .8868 & .1491 & 24.77 & .7778 & .1291 \\
			Cubic B-spline& 33.24 & .9395 & .0473 & \textbf{31.06} & \textbf{.8883} & \textbf{.1454} & \textbf{26.11} & \textbf{.8157} & \textbf{.1067} \\
			\hline
		\end{tabular}
	}
	\label{tab:ablation_trajectory}
}

\end{table}

\begin{table*}[!t]
	\centering
	\caption{Quantitative comparisons for \textbf{deblurring} on the synthetic dataset of Deblur-NeRF \cite{deblur-nerf}, referred to as DB-NeRF in the table due to space constraints. Notably, DB-NeRF* and DP-NeRF* are trained with ground-truth poses, while the others are trained with poses estimated by COLMAP \cite{colmap}. The experiments highlight the superior performance of our method over previous approaches. Additionally, the results demonstrate the sensitivity of Deblur-NeRF and DP-NeRF to pose accuracy. Regarding rendering efficiency, our method achieves over 200 FPS, whereas Deblur-NeRF, DP-NeRF, and BAD-NeRF fall below 1 FPS. Our method takes about 30 minutes to train, while other methods take more than 10 hours. Each color shading indicates the \colorbox{red!25}{best} and \colorbox{orange!25}{second-best} result, respectively.}
	\resizebox{\linewidth}{!}{
		\begin{tabular}{c|ccc|ccc|ccc|ccc|ccc}
			\hline
			& \multicolumn{3}{c|}{Cozyroom}& \multicolumn{3}{c|}{Factory} &  \multicolumn{3}{c|}{Pool} & \multicolumn{3}{c|}{Tanabata} & \multicolumn{3}{c}{Trolley} \\
			& {\scriptsize PSNR$\uparrow$} & {\scriptsize SSIM$\uparrow$} & {\scriptsize LPIPS$\downarrow$} & {\scriptsize PSNR$\uparrow$} & {\scriptsize SSIM$\uparrow$} & {\scriptsize LPIPS$\downarrow$} & {\scriptsize PSNR$\uparrow$} & {\scriptsize SSIM$\uparrow$} & {\scriptsize LPIPS$\downarrow$}  & {\scriptsize PSNR$\uparrow$} & {\scriptsize SSIM$\uparrow$} & {\scriptsize LPIPS$\downarrow$}  & {\scriptsize PSNR$\uparrow$} & {\scriptsize SSIM$\uparrow$} & {\scriptsize LPIPS$\downarrow$} \\
			\hline
			NeRF~\cite{nerf} & 26.13 & .7886 & .2484 & 22.02 & .5581 & .4550 & 29.90 & .7901 & .2595 & 20.57 & .5653 & .4584 & 21.71 & .6413 & .3814 \\
			3D-GS \cite{kerbl3Dgaussians} & 25.86 & .7908 & .2270 & 21.73 & .5623 & .4503 & 29.37 & .7816 & .2561 & 20.51 & .5773 & .4278 & 21.65 & .6587 & .3614 \\
			\hline
			MPR \cite{zamir2021multi} & 29.90 & .8862 & .0915 & 25.07 & .6994 & .2409 & \cellcolor{orange!25}33.28 & \cellcolor{orange!25}.8938 & .1290 & 22.60 & .7203 & .2507 & 26.24 & .8356 & .1762 \\
			SRN \cite{Tao2018CVPR} & 29.47 & .8759 & .0950 & 26.54 & .7604 & .2404 & 32.94 & .8847 & .1045 & 23.19 & .7274 & .2438 & 25.36 & .8119 & .1618 \\
			\hline
			DB-NeRF \cite{deblur-nerf}  & 29.53 & .8786 & .0879 & 25.85 & .7651 & .2340 & 31.07 & .8477 & .1397 & 23.20 & .7243 & .2471 & 25.68 & .8111 & .1683 \\
			DB-NeRF* & 30.26 & .8933 & .0791 & 26.40 & .7991 & .2191 & 32.30 & .8755 & .1345 & 24.56 & .7749 & .2166 & 26.24 & .8254 & .1671 \\
			DP-NeRF \cite{Lee_2023_CVPR} & 29.02 & .8761 & .0773 & 25.42 & .7559 & .2139 & 30.48 & .8402 & .1127 & 23.93 & .7540 & .2038 & 26.17 & .8269 & .1359 \\
			DP-NeRF* & 30.77 & .9020 & .0584 & 27.69 & .8328 & .1847 & 33.22 & .8922 & .0954 & 25.27 & .7973 & .1779 & 26.99 & .8413 & .1312 \\
			BAD-NeRF \cite{wang2023badnerf} & \cellcolor{orange!25}32.11 & \cellcolor{orange!25}.9137 & \cellcolor{orange!25}.0514 & \cellcolor{red!25}32.18 & \cellcolor{orange!25}.9105 & \cellcolor{orange!25}.1189 & 32.22 & .8680 & \cellcolor{orange!25}.0908 & \cellcolor{orange!25}25.80 & \cellcolor{orange!25}.8047 & \cellcolor{orange!25}.1563 & \cellcolor{orange!25}29.68 & \cellcolor{orange!25}.8954 & \cellcolor{orange!25}.0752 \\
			\hline
			Ours & \cellcolor{red!25}34.68 & \cellcolor{red!25}.9521 & \cellcolor{red!25}.0258 & \cellcolor{orange!25}31.88 & \cellcolor{red!25}.9270 & \cellcolor{red!25}.0952 & \cellcolor{red!25}36.95 & \cellcolor{red!25}.9434 & \cellcolor{red!25}.0225 & \cellcolor{red!25}32.12 & \cellcolor{red!25}.9481 & \cellcolor{red!25}.0464 & \cellcolor{red!25}33.97 & \cellcolor{red!25}.9628 & \cellcolor{red!25}.0209 \\
			\hline
		\end{tabular}
	}
	\label{tab:deblur_synthetic}
\end{table*}

\begin{table*}[!t]
	\centering
	\caption{Quantitative comparisons for \textbf{novel view synthesis} on the synthetic dataset of Deblur-NeRF \cite{deblur-nerf}, referred to as DB-NeRF in the table due to space constraints. The results demonstrate Our methods outperform previous state-of-the-art approaches, delivering the best performance across the board.}
	\resizebox{\linewidth}{!}{
		\begin{tabular}{c|ccc|ccc|ccc|ccc|ccc}
			\hline
			& \multicolumn{3}{c|}{Cozyroom}& \multicolumn{3}{c|}{Factory} &  \multicolumn{3}{c|}{Pool} & \multicolumn{3}{c|}{Tanabata} & \multicolumn{3}{c}{Trolley} \\
			& {\scriptsize PSNR$\uparrow$} & {\scriptsize SSIM$\uparrow$} & {\scriptsize LPIPS$\downarrow$} & {\scriptsize PSNR$\uparrow$} & {\scriptsize SSIM$\uparrow$} & {\scriptsize LPIPS$\downarrow$} & {\scriptsize PSNR$\uparrow$} & {\scriptsize SSIM$\uparrow$} & {\scriptsize LPIPS$\downarrow$}  & {\scriptsize PSNR$\uparrow$} & {\scriptsize SSIM$\uparrow$} & {\scriptsize LPIPS$\downarrow$}  & {\scriptsize PSNR$\uparrow$} & {\scriptsize SSIM$\uparrow$} & {\scriptsize LPIPS$\downarrow$} \\
			\hline
			NeRF~\cite{nerf} & 25.45 & .7734 & .2514 & 22.51 & .5672 & .4199 & 29.23 & .7711 & .2756 & 20.74 & .5774 & .4341 & 21.09 & .6157 & .3924 \\
			3D-GS \cite{kerbl3Dgaussians} & 25.24 & .7789 & .2084 & 21.34 & .5532 & .4371 & 28.61 & .7562 & .2610 & 20.37 & .5755 & .3958 & 20.86 & .6167 & .3701 \\
			\hline
			MPR \cite{zamir2021multi}+\cite{kerbl3Dgaussians} & 29.00 & .8685 & .0894 & 23.38 & .6848 & .2374 & 29.04 & .8170 & .1761 & 22.53 & .7349 & .2162 & 26.27 & .8352 & .1694 \\
			SRN \cite{Tao2018CVPR}+\cite{kerbl3Dgaussians} & 28.64 & .8662 & .0903 & 25.45 & .7740 & .2045 & 29.97 & .8207 & .1650 & 23.09 & .7456 & .2141 & 24.89 & .8103 & .1666 \\
			\hline
			DB-NeRF \cite{deblur-nerf}  & 29.09 & .8718 & .0937 & 25.19 & .7359 & .2315 & 30.97 & .8401 & .1600 & 23.71 & .7483 & .2369 & 24.46 & .7758 & .2215 \\
			DB-NeRF* & 29.88 & .8901 & .0747 & 26.06 & .8023 & .2106 & 30.94 & .8399 & .1694 & 24.82 & .7861 & .2045 & 25.78 & .8122 & .1797 \\
			DP-NeRF \cite{Lee_2023_CVPR} & 29.56 & .8831 & .0683 & 27.91 & .8317 & .1833 & 30.98 & .8454 & .1265 & 24.73 & .7867 & .1865 & 25.62 & .8123 & .1606 \\
			DP-NeRF* & 30.16 & .8958 & .0608 & \cellcolor{orange!25}28.43 & .8420 & .1771 & \cellcolor{orange!25}31.78 & \cellcolor{orange!25}.8623 & .1306 & 25.37 & \cellcolor{orange!25}.8046 & .1701 & 26.44 & .8307 & .1424 \\
			BAD-NeRF \cite{wang2023badnerf} & \cellcolor{orange!25}31.07 & \cellcolor{orange!25}.9027 & \cellcolor{orange!25}.0551 & \cellcolor{red!25}31.71 & \cellcolor{orange!25}.9038 & \cellcolor{orange!25}.1204 & 30.95 & .8389 & \cellcolor{orange!25}.1107 & \cellcolor{orange!25}25.42 & .8011 & \cellcolor{orange!25}.1566 & \cellcolor{orange!25}28.60 & \cellcolor{orange!25}.8806 & \cellcolor{orange!25}.0843 \\
			\hline
			Ours & \cellcolor{red!25}32.35 & \cellcolor{red!25}.9278 & \cellcolor{red!25}.0340 & 28.18 & \cellcolor{red!25}.9164 & \cellcolor{red!25}.0999 & \cellcolor{red!25}33.30 & \cellcolor{red!25}.8964 & \cellcolor{red!25}.0551 & \cellcolor{red!25}31.00 & \cellcolor{red!25}.9427 & \cellcolor{red!25}.0478 & \cellcolor{red!25}31.44 & \cellcolor{red!25}.9443 & \cellcolor{red!25}.0345 \\
			\hline
		\end{tabular}
	}
	\label{tab:nvs_synthetic}
\end{table*}

\begin{table*}[!t]
	\centering
	\caption{Quantitative comparisons for \textbf{deblurring} on the synthetic dataset of MBA-VO \cite{deblur-nerf}. The results demonstrate that our method achieves the best performance even when the camera undergoes acceleration}
	\resizebox{0.9\linewidth}{!}{
		\begin{tabular}{c|ccc|ccc|ccc}
			\hline
			& \multicolumn{3}{c|}{ArchViz-low}& \multicolumn{3}{c|}{ArchViz-high} &  \multicolumn{3}{c}{Average} \\
			& {\scriptsize PSNR$\uparrow$} & {\scriptsize SSIM$\uparrow$} & {\scriptsize LPIPS$\downarrow$} & {\scriptsize PSNR$\uparrow$} & {\scriptsize SSIM$\uparrow$} & {\scriptsize LPIPS$\downarrow$} & {\scriptsize PSNR$\uparrow$} & {\scriptsize SSIM$\uparrow$} & {\scriptsize LPIPS$\downarrow$} \\
			\hline
			NeRF~\cite{nerf}& 26.26 & .7887 & .3775 & 23.29 & .7098 & .4910 & 24.78 & .7493 & .4343 \\
			3D-GS \cite{kerbl3Dgaussians}& 26.54 & .8113 & .3385 & 23.43 & .7243 & .4614 & 24.99 & .7678 & .4000 \\
			\hline
			MPR \cite{zamir2021multi} & 29.60 & .8757 & .2103 & 25.04 & .7711 & .3576 & 27.32 & .8234 & .2840 \\
			SRN \cite{Tao2018CVPR} & 30.15 & \cellcolor{orange!25}.8814 & .1703 & 27.07 & \cellcolor{orange!25}.8190 & .2796 & 28.61 & \cellcolor{orange!25}.8502 & .2250 \\
			\hline
			DB-NeRF \cite{deblur-nerf}  & 28.38 & .8484 & .1792 & 25.71 & .7762 & .3220 & 27.05 & .8123 & .2506 \\
			DB-NeRF*& 29.65 & .8744 & .1764 & 26.44 & .8010 & .3172 & 28.05 & .8377 & .2468 \\
			DP-NeRF \cite{Lee_2023_CVPR} & 28.24 & .8490 & .1615 & 25.74 & .7839 & .2913 & 26.99 & .8165 & .2264 \\
			DP-NeRF* & 28.57 & .8481 & \cellcolor{orange!25}.1511 & 26.80 & .8081 & .2736 & 27.69 & .8281 & \cellcolor{orange!25}.2124 \\
			BAD-NeRF \cite{wang2023badnerf}& \cellcolor{orange!25}30.51 & .8749 & .1654 & \cellcolor{orange!25}27.54 & .8109 & \cellcolor{orange!25}.2679 & \cellcolor{orange!25}29.03 & .8429 & .2167 \\
			\hline
			Ours & \cellcolor{red!25}32.28   & \cellcolor{red!25}.9167 & \cellcolor{red!25}.1134  & \cellcolor{red!25}29.64 & \cellcolor{red!25}.8568  & \cellcolor{red!25}.1847  & \cellcolor{red!25}30.96  & \cellcolor{red!25}.8868  & \cellcolor{red!25}.1491  \\
			\hline
		\end{tabular}
	}
	\label{tab:deblur_synthetic_room}
\end{table*}

\subsection{Experimental Settings}
\PAR{Implementation Details.} We implemented our method using PyTorch \cite{paszke2019pytorch} within the 3D-GS \cite{kerbl3Dgaussians} framework. Both the optimization of Gaussians and camera poses are performed using the Adam optimizer. The learning rate for Gaussians remains identical to the original 3D-GS \cite{kerbl3Dgaussians}, while for camera poses (i.e., $\bT_\mathrm{start}$ and $\bT_\mathrm{end}$ in Equation \eqref{eq_trajectory}), it is exponentially decreased from $1 \times 10^{-3}$ to $1 \times 10^{-5}$. The number of virtual camera poses (i.e., $n$ in Equation \eqref{eq_blur_im_formation}) is set to 10, considering the trade-off between performance and efficiency. We initialize camera poses and Gaussians using estimations obtained from COLMAP \cite{colmap}. All experiments are conducted on an NVIDIA RTX 4090 GPU.

\PAR{Benchmark Datasets.}We evaluate the performance of our method using both synthetic and real datasets provided by Deblur-NeRF \cite{deblur-nerf} and BAD-NeRF \cite{wang2023badnerf}.
Deblur-NeRF \cite{wang2023badnerf} contains five scenes, where the blurred images are generated via Blender \cite{Blender} by averaging sharp virtual images captured during the exposure time, under the assumption of consistent velocity camera motion.
For our evaluation, we utilize the dataset from BAD-NeRF \cite{wang2023badnerf}, which expands upon the number of virtual sharp images to create more realistic motion-blurred images while keeping other settings consistent with \cite{deblur-nerf}.
Additionally, Deblur-NeRF \cite{deblur-nerf} captured a real motion-blurred dataset by intentionally shaking a handheld camera during exposure.

To enhance the evaluation of our method on severely motion-blurred images, we incorporate a dataset tailored for motion blur-aware visual odometry benchmarking (\ie MBA-VO \cite{mba-vo}).
Different from the assumption of constant velocity in the Deblur-NeRF dataset \cite{deblur-nerf}, the blur images synthesized from MBA-VO \cite{mba-vo} are based on real camera motion trajectories obtained from the ETH3D dataset \cite{schops2019bad}, which do not exhibit constant velocity and include accelerations, thus presenting notable challenges.

\PAR{Baselines and Evaluation Metrics.} We conduct a comparative \textbf{deblurring} analysis between our method and state-of-the-art learning-based single image deblurring algorithms including SRN \cite{Tao2018CVPR} and MPR \cite{zamir2021multi}. Additionally, we include evaluations against closely related approaches such as Deblur-NeRF \cite{deblur-nerf}, DP-NeRF \cite{Lee_2023_CVPR}, and BAD-NeRF \cite{wang2023badnerf}. To evaluate deblurring performance, we render high-quality images corresponding to the midpoint (\ie $\bC_\mathrm{\frac{\tau}{2}}$ in \eqnref{eq_continuous_blur_im_formation}) of exposure time of each training image from the optimized Gaussians.

To facilitate \textbf{novel view synthesis} using SRN \cite{Tao2018CVPR} and MPR \cite{zamir2021multi}, we use the deblurred images obtained from pre-trained models as inputs for training 3D-GS, as these single deblurring methods are not optimized for synthesizing novel images. During the training stage, all poses are estimated using COLMAP \cite{colmap}. Additionally, we train Deblur-NeRF \cite{deblur-nerf} and DP-NeRF \cite{Lee_2023_CVPR} with ground-truth camera poses from Blender to assess the impact of inaccurate pose estimation on these two methods. The quality of the rendered sharp image is evaluated using PSNR, SSIM, and LPIPS metrics.

For \textbf{pose accuracy} evaluation, we compare the Absolute Trajectory Error (ATE) metric, against classical structure-from-motion framework COLMAP \cite{colmap} and BAD-NeRF \cite{wang2023badnerf}.

\begin{figure*}[t]
    \setlength\tabcolsep{1pt}
    \centering
    \begin{tabular}{cccccccc}
        &\raisebox{-0.02in}{\rotatebox[origin=t]{90}{\tiny{3D-GS}}} &
        \includegraphics[valign=m,width=0.98\textwidth]{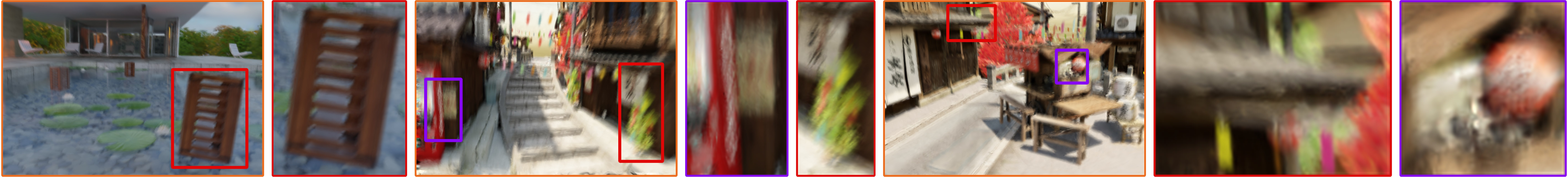}\\
        \specialrule{0em}{.05em}{.05em}
        \raisebox{-0.035in}{\rotatebox[origin=t]{90}{\tiny{SRN\cite{Tao2018CVPR}}}}
        &\raisebox{-0.035in}{\rotatebox[origin=t]{90}{\tiny{+3D-GS}}} &
        \includegraphics[valign=m,width=0.98\textwidth]{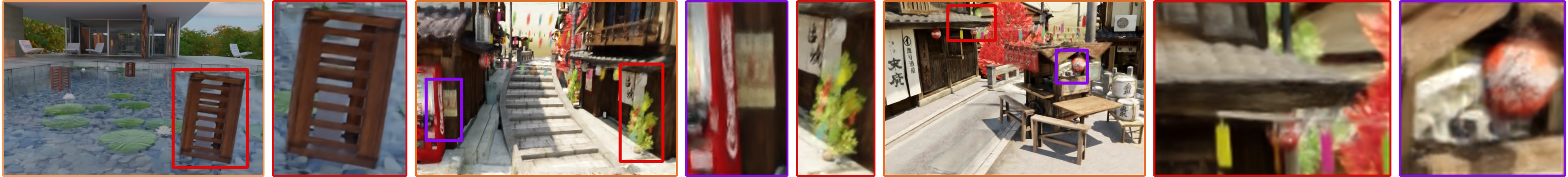}\\
        \specialrule{0em}{.05em}{.05em}
        \raisebox{-0.035in}{\rotatebox[origin=t]{90}{\tiny{Deblur-}}}
        &\raisebox{-0.035in}{\rotatebox[origin=t]{90}{\tiny{NeRF*\cite{deblur-nerf}}}} &
        \includegraphics[valign=m,width=0.98\textwidth]{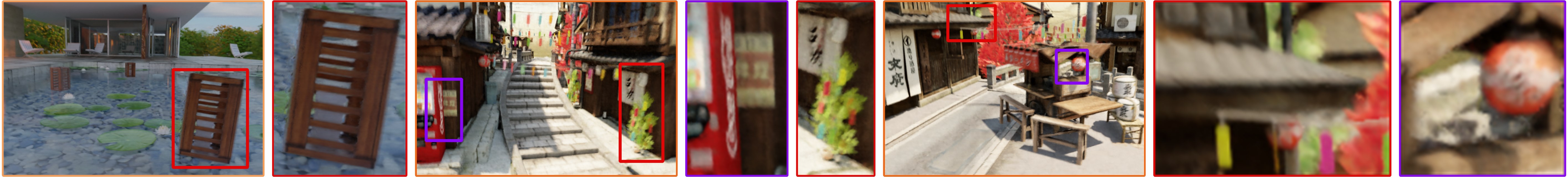}\\
        \specialrule{0em}{.05em}{.05em}
        \raisebox{-0.035in}{\rotatebox[origin=t]{90}{\tiny{DP-}}}
        &\raisebox{-0.035in}{\rotatebox[origin=t]{90}{\tiny{NeRF*\cite{Lee_2023_CVPR}}}} &
        \includegraphics[valign=m,width=0.98\textwidth]{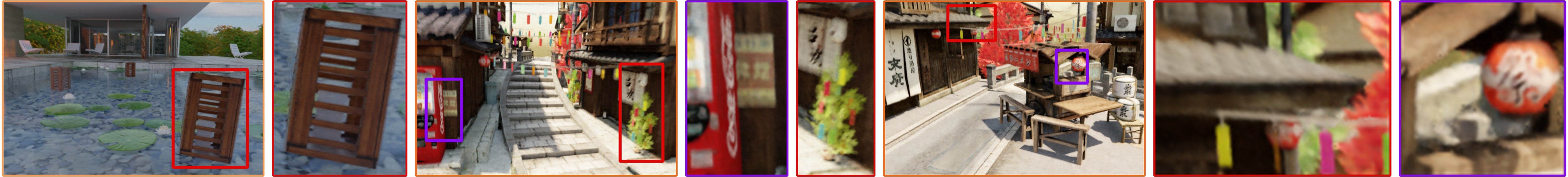}\\
        \specialrule{0em}{.05em}{.05em}
        \raisebox{-0.035in}{\rotatebox[origin=t]{90}{\tiny{BAD-}}}
        &\raisebox{-0.035in}{\rotatebox[origin=t]{90}{\tiny{NeRF\cite{wang2023badnerf}}}} &
        \includegraphics[valign=m,width=0.98\textwidth]{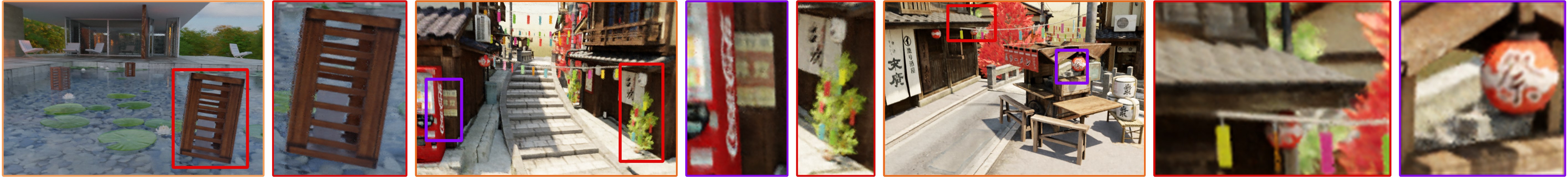}\\
        \specialrule{0em}{.05em}{.05em}
        &\raisebox{-0.035in}{\rotatebox[origin=t]{90}{\scriptsize Ours}} &
        \includegraphics[valign=m,width=0.98\textwidth]{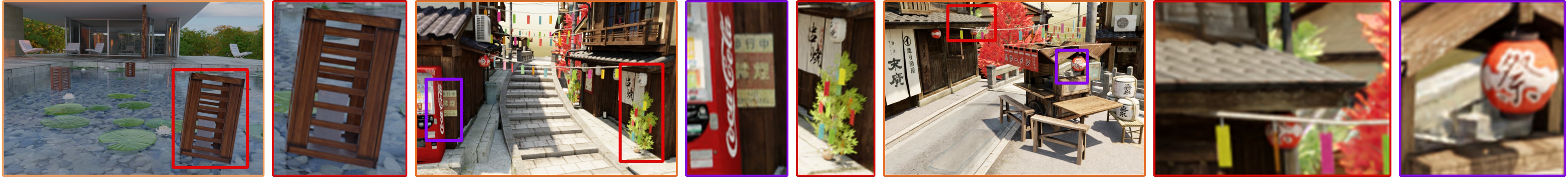}\\
        \specialrule{0em}{.05em}{.05em}
        &\raisebox{-0.035in}{\rotatebox[origin=t]{90}{\tiny{Reference}}} &
        \includegraphics[valign=m,width=0.98\textwidth]{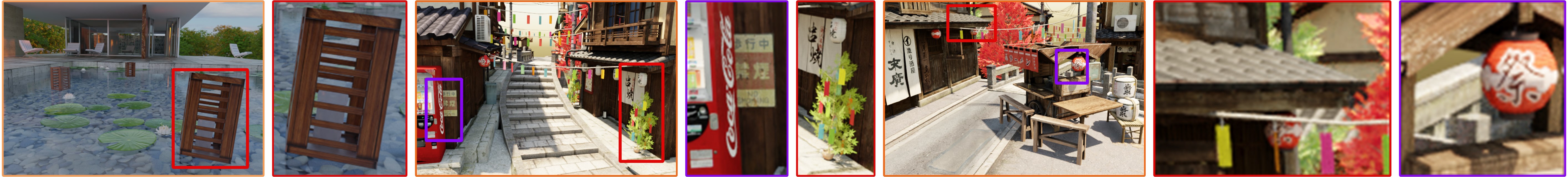}\\
    \end{tabular}
    \vspace{0.5em}
    \caption{ {\textbf{Qualitative novel view synthesis results of different methods with synthetic datasets.} Despite being trained with ground truth poses (*), BAD-Gaussians outperforms Deblur-NeRF* and DP-NeRF* in recovering high-quality scenes from motion-blurred images with inaccurate camera poses, showcasing its superior performance.}}
    \label{fig:comparison-syn}
    \vspace{-1em}
\end{figure*}

\subsection{Ablation Study}
\PAR{Virtual Camera Poses.} We experimented to investigate the effect of the number of interpolated virtual camera poses within the exposure time, as described by $n$ in \eqnref{eq_blur_im_formation}. Two of the five synthetic scenes provided by Deblur-NeRF \cite{deblur-nerf} were selected, representing sequences with minor and severe motion blur (\ie \textit{Cozy2room} and \textit{Tanabata}), respectively. We varied the number, $n$, from 3 to 20 in our study, and the rendering metric results are presented in Table \ref{tab:ablation_n}. Our experimental results demonstrate that increasing the number of interpolated virtual camera poses helps in addressing severe motion blur. However, marginal improvements are observed for minor blur instances. Based on our experiments, we choose $n=10$ interpolated virtual images to strike a balance between rendering performance and training efficiency (larger $n$ means more computational resources).

\PAR{Trajectory Representations.} To assess the impact of different trajectory representations, we conduct two experiments: one involves optimizing $\mathbf{T}_\mathrm{start}$ and $\mathbf{T}_\mathrm{end}$ to depict a linear trajectory, while the other utilizes a higher-order spline (i.e., cubic B-spline) that jointly optimizes four control knots $\mathbf{T}_\mathrm{1}$, $\mathbf{T}_\mathrm{2}$, $\mathbf{T}_\mathrm{3}$, and $\mathbf{T}_\mathrm{4}$ to capture more complex camera motions. We refer \cite{wangsupplementary} for more details about cubic B-spline. The average quantitative results on the synthetic datasets of \textit{Deblur-NeRF-S} \cite{deblur-nerf} (Cozy2room, Factory, Pool, Tanabata and Trolley) and \textit{MBA-VO} \cite{mba-vo} (ArchViz-low and ArchViz-high), along with real datasets, \textit{Deblur-NeRF-S}, which includes 10 real captured scenes from Deblur-NeRF, are presented in Table \ref{tab:ablation_trajectory}. In synthetic scenes, linear interpolation demonstrates comparable performance to cubic B-spline interpolation. However, in real captured scenes, cubic B-spline interpolation outperforms linear interpolation, particularly due to the longer exposure time during image capture. The effectiveness of cubic B-spline interpolation in real scenes can be attributed to its ability to better model the nuances of camera motion over longer time intervals. Conversely, linear interpolation is sufficient to accurately represent camera motion trajectories within shorter time intervals, as observed in synthetic scenes. Combining the training efficiency and rendering quality, we employ linear interpolation in synthetic datasets and cubic B-spline in real data. 

\begin{table*}[!t]
	\centering
	\caption{Quantitative comparisons for \textbf{novel view synthesis} on the real captured dataset of Deblur-NeRF \cite{deblur-nerf}.}
	\resizebox{\linewidth}{!}{
		\begin{tabular}{c|ccc|ccc|ccc|ccc|ccc}
			\hline
			& \multicolumn{3}{c|}{Ball}& \multicolumn{3}{c|}{Basket} &  \multicolumn{3}{c|}{Buick} & \multicolumn{3}{c|}{Coffee} & \multicolumn{3}{c}{Decoration} \\
			& {\scriptsize PSNR$\uparrow$} & {\scriptsize SSIM$\uparrow$} & {\scriptsize LPIPS$\downarrow$} & {\scriptsize PSNR$\uparrow$} & {\scriptsize SSIM$\uparrow$} & {\scriptsize LPIPS$\downarrow$} & {\scriptsize PSNR$\uparrow$} & {\scriptsize SSIM$\uparrow$} & {\scriptsize LPIPS$\downarrow$}  & {\scriptsize PSNR$\uparrow$} & {\scriptsize SSIM$\uparrow$} & {\scriptsize LPIPS$\downarrow$}  & {\scriptsize PSNR$\uparrow$} & {\scriptsize SSIM$\uparrow$} & {\scriptsize LPIPS$\downarrow$} \\
			\hline
			NeRF \cite{nerf} & 24.08 & .6237 & .3992 & 23.72 & .7086 & .3223 & 21.59 & .6325 & .3502 & 26.48 & .8064 & .2896 & 22.39 & .6609 & .3633 \\
			3D-GS \cite{kerbl3Dgaussians} & 23.72 & .6321 & .3210 & 23.96 & .7466 & .2600 & 21.53 & .6630 & .2743 & 27.44 & .6630 & .2208 & 22.29 & .6826 & .2868 \\
			\hline
			DB-NeRF \cite{deblur-nerf} & \cellcolor{orange!25}27.15 & \cellcolor{orange!25}.7641 & .2112 & 27.35 & .8367 & .1347 & \cellcolor{orange!25}24.93 & .7791 & .1545 & 30.72 & .8949 & .1070 & 24.15 & .7730 & .1700 \\
			DP-NeRF \cite{Lee_2023_CVPR} & 26.86 & .7522 & \cellcolor{orange!25}.2066 &\cellcolor{red!25}27.71 & \cellcolor{orange!25}.8434 & .1301 & \cellcolor{red!25}25.52 & \cellcolor{orange!25}.7847 & \cellcolor{orange!25}.1433 & \cellcolor{orange!25}31.35 & \cellcolor{orange!25}.9013 & \cellcolor{orange!25}.0987 & \cellcolor{orange!25}24.42 & \cellcolor{orange!25}.7820 & \cellcolor{orange!25}.1611 \\
			BAD-NeRF \cite{wang2023badnerf} & 26.71 & .7480 & .2120 & 26.40 & .8024 & \cellcolor{orange!25}.1268 & 23.06 & .7104 & .2099 & 29.08 & .8401 & .1941 & 22.09 & .6067 & .3079 \\
			\hline
			Ours & \cellcolor{red!25}28.11 & \cellcolor{red!25}.8041 & \cellcolor{red!25}.2058 & \cellcolor{orange!25}27.41 & \cellcolor{red!25}.8632 & \cellcolor{red!25}.0987 & 24.22 & \cellcolor{red!25}.8064 & \cellcolor{red!25}.1161 & \cellcolor{red!25}32.17 & \cellcolor{red!25}.9280 & \cellcolor{red!25}.0659 & \cellcolor{red!25}25.53 & \cellcolor{red!25}.8442 & \cellcolor{red!25}.0937 \\
			\hline
			\hline
			& \multicolumn{3}{c|}{Girl}& \multicolumn{3}{c|}{Heron} &  \multicolumn{3}{c|}{Parterre} & \multicolumn{3}{c|}{Puppet} & \multicolumn{3}{c}{Stair} \\
			& {\scriptsize PSNR$\uparrow$} & {\scriptsize SSIM$\uparrow$} & {\scriptsize LPIPS$\downarrow$} & {\scriptsize PSNR$\uparrow$} & {\scriptsize SSIM$\uparrow$} & {\scriptsize LPIPS$\downarrow$} & {\scriptsize PSNR$\uparrow$} & {\scriptsize SSIM$\uparrow$} & {\scriptsize LPIPS$\downarrow$}  & {\scriptsize PSNR$\uparrow$} & {\scriptsize SSIM$\uparrow$} & {\scriptsize LPIPS$\downarrow$}  & {\scriptsize PSNR$\uparrow$} & {\scriptsize SSIM$\uparrow$} & {\scriptsize LPIPS$\downarrow$} \\
			\hline
			NeRF \cite{nerf} & 20.07 & .7075 & .3196 & 20.50 & .5217 & .4129 & 23.14 & .6201 & .4046 & 22.09 & .6093 & .3389 & 22.87 & .4561 & .4868 \\
			3D-GS \cite{kerbl3Dgaussians} & 19.97 & .7276 & .2613 & 20.28 & .5254 & .3109 & 22.98 & .6326 & .2967 & 22.38 & .6463 & .2645 & 22.68 & .4709 & .3911 \\
			\hline
			DB-NeRF \cite{deblur-nerf}& \cellcolor{orange!25}22.27 & .7976 & .1687 & 22.63 & .6874 & .2099 & 25.82 & .7597 & .2161 & 25.24 & .7510 & .1577 & 25.39 & .6296 & .2102 \\
			DP-NeRF \cite{Lee_2023_CVPR} & \cellcolor{red!25}23.43 & \cellcolor{orange!25}.8148 & \cellcolor{orange!25}.1459 & \cellcolor{orange!25}22.79 & \cellcolor{orange!25}.7010 & \cellcolor{orange!25}.1891 & \cellcolor{orange!25}25.90 & \cellcolor{orange!25}.7658 & \cellcolor{orange!25}.1893 & \cellcolor{red!25}25.56 & \cellcolor{orange!25}.7560 & \cellcolor{orange!25}.1469 & 25.53 & .6326 & .1778 \\
			BAD-NeRF \cite{wang2023badnerf} & 19.72 & .6194 & .3378 & 21.81 & .6249 & .2340 & 24.86 & .7066 & .2131 & 24.14 & .7073 & .1833 & \cellcolor{orange!25}25.64 & \cellcolor{orange!25}.6370 & \cellcolor{orange!25}.1768 \\
			\hline
			Ours & 21.28 & \cellcolor{red!25}.8152 & \cellcolor{red!25}.1040 & \cellcolor{red!25}24.52 & \cellcolor{red!25}.7657 & \cellcolor{red!25}.1187 & \cellcolor{red!25}25.94 & \cellcolor{red!25}.8133 & \cellcolor{red!25}.0983 & \cellcolor{orange!25}25.25 & \cellcolor{red!25}.7991 & \cellcolor{red!25}.0948 & \cellcolor{red!25}26.63 & \cellcolor{red!25}.7177 & \cellcolor{red!25}.0685 \\
			\hline
		\end{tabular}
	}
 	\label{tab:nvs_real}
	\vspace{-1em}
\end{table*}

\begin{figure*}[!t]
	\setlength\tabcolsep{1pt}
	\centering
	\begin{tabular}{cccccc}
		\includegraphics[width=0.16\textwidth]{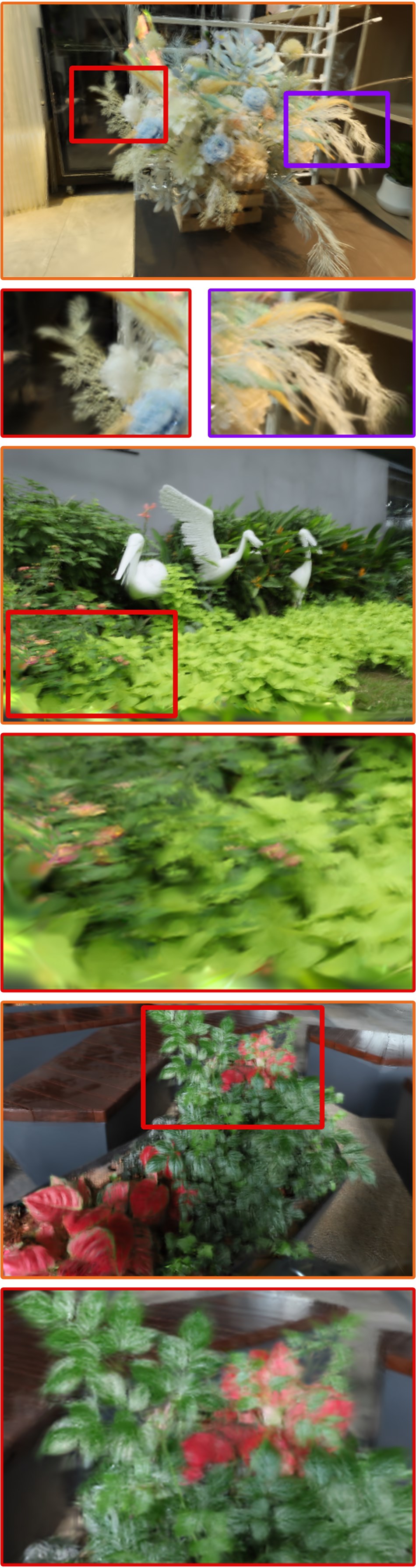} &
		\includegraphics[width=0.16\textwidth]{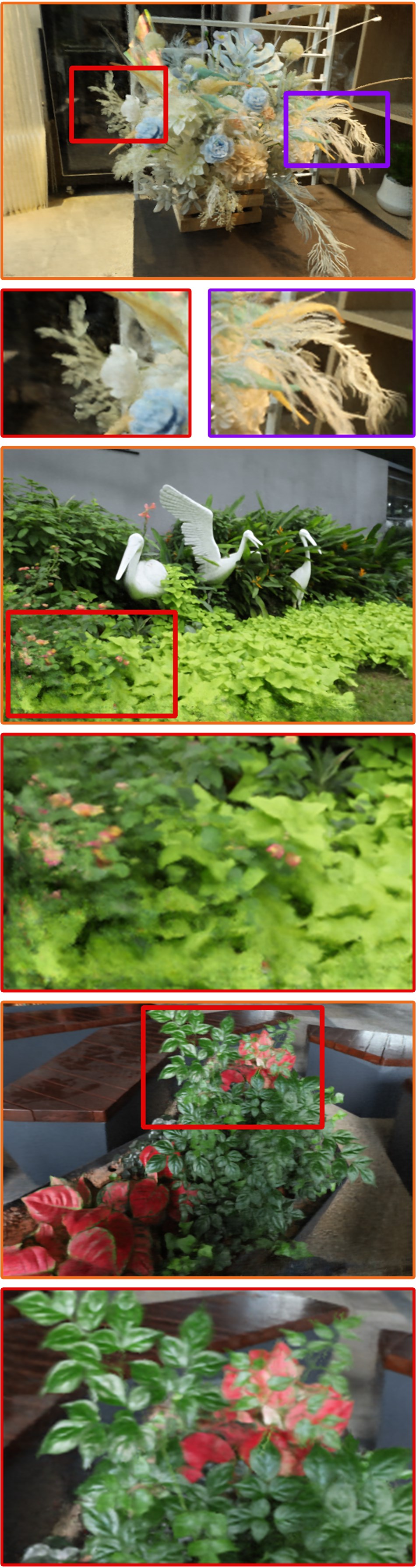} &
		\includegraphics[width=0.16\textwidth]{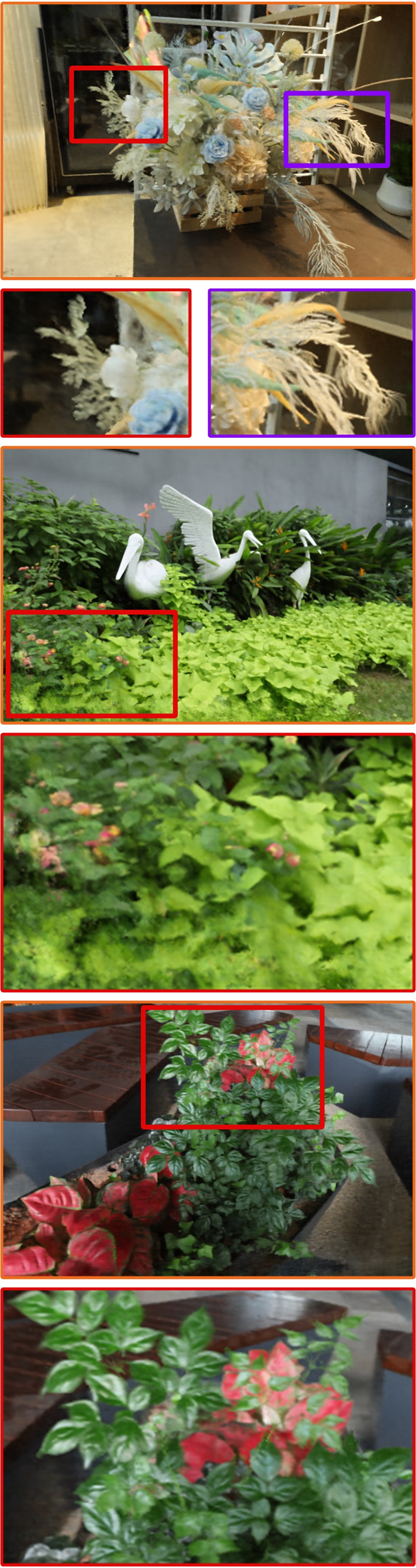} &
		\includegraphics[width=0.16\textwidth]{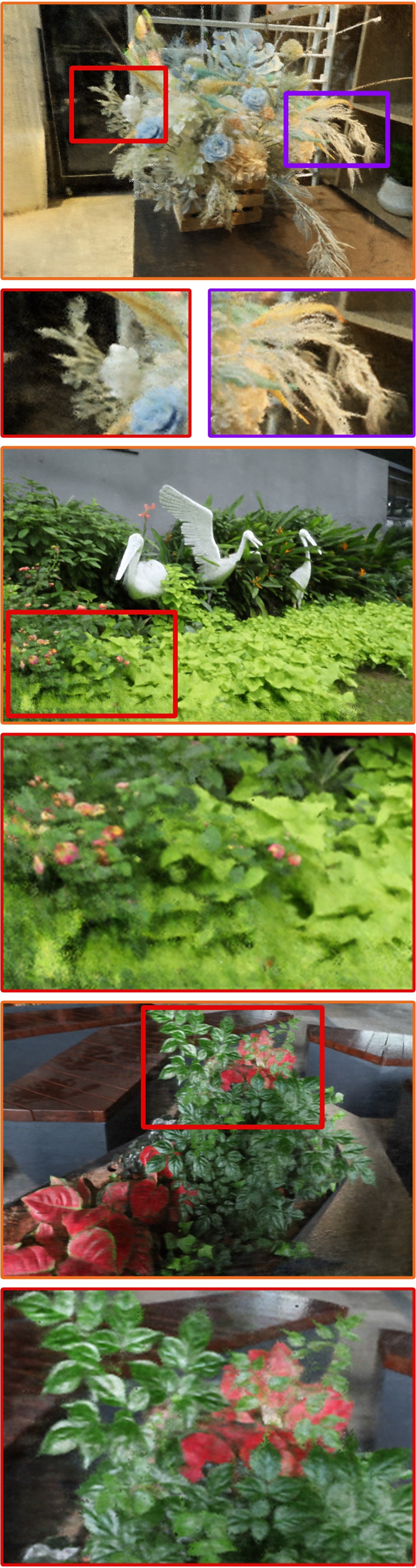} &
		\includegraphics[width=0.16\textwidth]{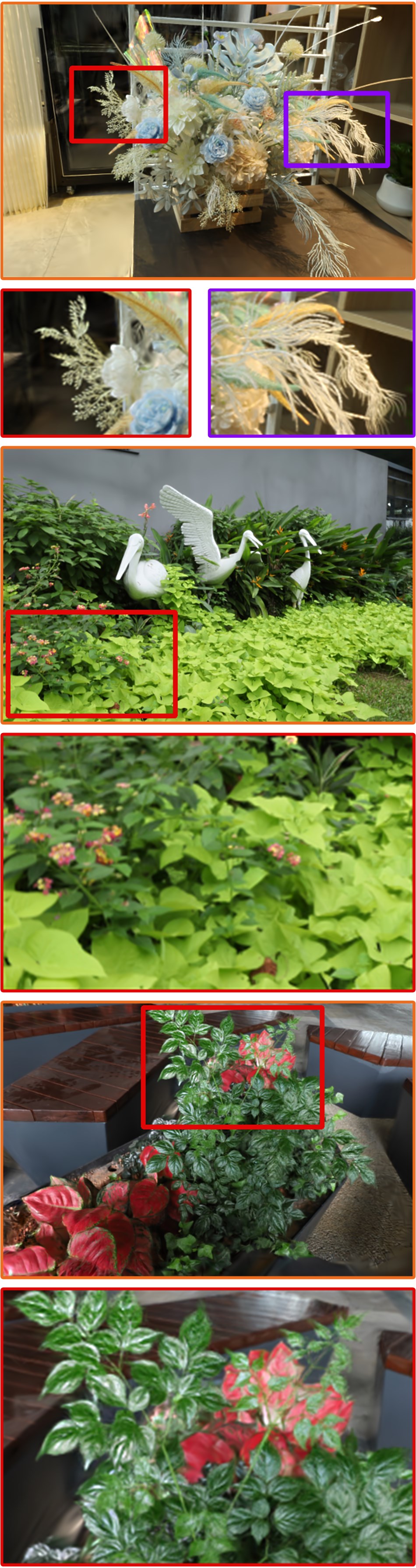} &
		\includegraphics[width=0.16\textwidth]{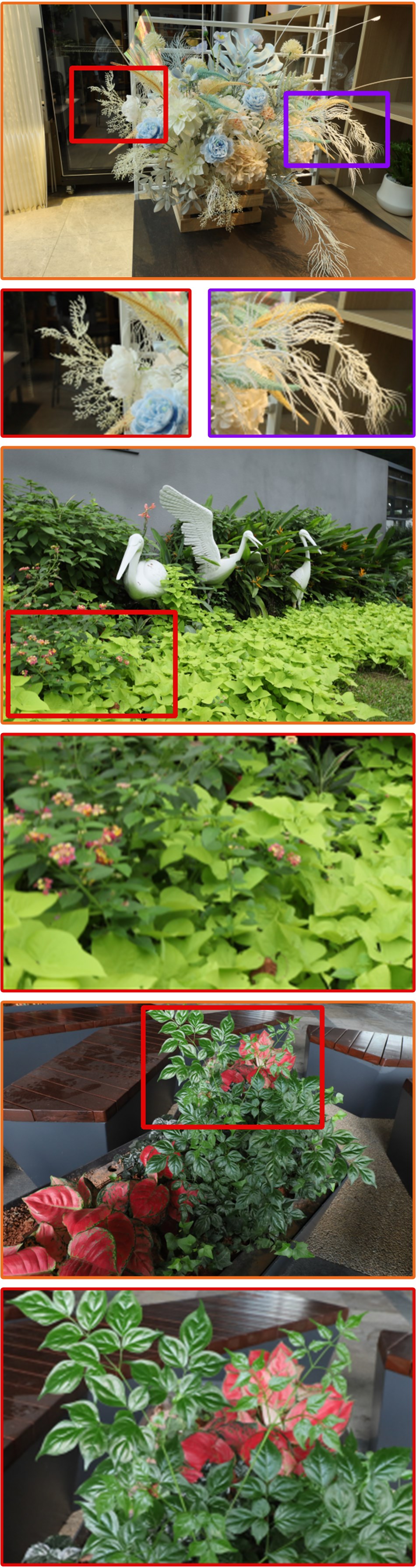} \\
		\specialrule{0em}{0.05pt}{0.05pt}
		\tiny{3D-GS\cite{kerbl3Dgaussians}} & \tiny{Deblur-NeRF\cite{deblur-nerf}} & \tiny{DP-NeRF\cite{Lee_2023_CVPR}} & \tiny{BAD-NeRF\cite{wang2023badnerf}} & \tiny{Ours} & \tiny{Reference}
	\end{tabular}
	\caption{{\textbf{Qualitative novel view synthesis results of different methods with the real datasets.}} The experimental results demonstrate that our method achieves superior performance over prior methods on the real dataset as well. 
In contrast, BAD-NeRF yields poorer results when applied to real data and exhibits satisfactory performance only within synthetic datasets.}
	\label{fig:comparison-real}
\end{figure*}

\begin{table}
	\caption{{\textbf{Pose estimation performance of BAD-Gaussians on various blur sequences.}} The results are in the absolute trajectory error metric in centimeters (ATE/cm). The COLMAP-sharp / COLMAP-blur represents the result of COLMAP with sharp/blurry images respectively.}
    \setlength\tabcolsep{1pt}
    \centering
    \resizebox{\linewidth}{!}{
       \begin{tabular}{c|cccccccc}
        \specialrule{0.1em}{1pt}{1pt}
            & \footnotesize Cozy2room	& \footnotesize Factory	& \footnotesize Pool & \footnotesize Tanabata	& \footnotesize Trolley & \footnotesize ArchViz-low & \footnotesize ArchViz-high\\
        \hline
        \scriptsize {COLMAP-sharp\cite{colmap}}
            & \footnotesize.133$\pm$.079 & \footnotesize.101$\pm$.052 & \footnotesize.172$\pm$.047 & \footnotesize.277$\pm$.103 & \footnotesize.173$\pm$.079 & \footnotesize.148$\pm$.078 & \footnotesize.115$\pm$.066\\
        \hline
        \scriptsize {COLMAP-blur\cite{colmap}}
            & \cellcolor{orange!25}\footnotesize.287$\pm$.161 & \footnotesize1.69$\pm$1.24 & \footnotesize.633$\pm$.236 & \footnotesize1.59$\pm$1.06 & \footnotesize.912$\pm$.559 & \footnotesize.536$\pm$.309 & \cellcolor{red!25}\footnotesize.766$\pm$.391\\
        \scriptsize {BAD-NeRF\cite{wang2023badnerf}}
            & \footnotesize.316$\pm$.101 & \cellcolor{orange!25}\footnotesize.600$\pm$.208 & \cellcolor{red!25}\footnotesize.095$\pm$.046 & \cellcolor{orange!25}\footnotesize1.34$\pm$.543 & \cellcolor{red!25}\footnotesize.122$\pm$.070 & \footnotesize1.24$\pm$.635 & \footnotesize2.18$\pm$1.54\\
        \hline
        \scriptsize {Ours}
            & \footnotesize\cellcolor{red!25}.165$\pm$.057 & \footnotesize\cellcolor{red!25}.395$\pm$.426 & \cellcolor{orange!25}\footnotesize.264$\pm$.098 & \cellcolor{red!25}\footnotesize1.20$\pm$.621 & \cellcolor{orange!25}\footnotesize.134$\pm$.070 & \cellcolor{red!25}\footnotesize.520$\pm$.591 & \cellcolor{orange!25}\footnotesize1.41$\pm$1.11\\
        \specialrule{0.1em}{1pt}{1pt}
    \end{tabular}
    }
	\label{table_ate}
	\vspace{-1em}
\end{table}

\subsection{Results}

\PAR{Results on Synthetic Data.} We evaluate our approach against baseline methods using synthetic datasets obtained from both Deblur-NeRF \cite{deblur-nerf} and MBA-VO \cite{mba-vo}. The quantitative evaluation results on \textbf{deblurring} and \textbf{novel view synthesis} with the Deblur-NeRF dataset are presented in Table \ref{tab:deblur_synthetic} and Table \ref{tab:nvs_synthetic} respectively. All metrics (\ie PSNR, SSIM, LPIPS) demonstrate substantial improvements over prior state-of-the-art methods (\ie 3.6 and 1.7 dB higher than the second best method on average in Table \ref{tab:deblur_synthetic} and Table \ref{tab:nvs_synthetic}, respectively). 

The results demonstrate that both NeRF \cite{nerf} and 3D-GS \cite{kerbl3Dgaussians} are suffering from motion-blurred images, which motivates the necessity of our method. Single-stage methods such as MPR \cite{zamir2021multi} and SRN \cite{Tao2018CVPR} fall short of matching the performance of our approach due to their limited utilization of geometric information across multi-view images. Additionally, our method surpasses Deblur-NeRF \cite{deblur-nerf} and DP-NeRF \cite{Lee_2023_CVPR}, partly because they fail to optimize the inaccurate camera poses estimated from COLMAP during training. We additionally trained Deblur-NeRF and DP-NeRF with ground truth poses (denoted as *), and the results demonstrate improved performance, reflecting the sensitivity to the pose accuracies of both Deblur-NeRF and DP-NeRF. A notable result from both  Table \ref{tab:deblur_synthetic} and Table \ref{tab:nvs_synthetic} is that our method performs worse than BAD-NeRF on the Factory sequence. We find that it is caused by the inferior capability to represent the sky by Gaussian splatting compared to NeRF.  Nevertheless, our method still performs better than all prior methods with a large margin on average. 

Qualitative results are presented in Fig. \ref{fig:comparison-syn}. It demonstrates that our method effectively attains high-quality scene representations with intricate details, being trained from a series of blurred images. In Fig. \ref{fig:comparison-syn}, it is evident that Deblur-NeRF \cite{deblur-nerf} and DP-NeRF \cite{Lee_2023_CVPR} encounter difficulties in modeling regions with significant depth variations, attributable to their method of synthesizing blurred images through convolution with a blur kernel applied to rendered images. The physically based method BAD-NeRF \cite{wang2023badnerf} exhibits superior performance overall, yet it still presents some deficiencies, particularly noticeable around areas with significant color and depth variation. It demonstrates the effectiveness of the explicit Gaussian splatting representation over the implicit neural representation. 

We further assess the performance of our methods using a dataset characterized by severe motion blur and camera movements with varying velocities, sourced from MBA-VO \cite{mba-vo}. The results presented in Table \ref{tab:deblur_synthetic_room} demonstrate that our method achieves the best performance even when the camera undergoes acceleration.

\PAR{Results on Real Data.} The performance of \textbf{novel view synthesis} on real data sourced from Deblur-NeRF \cite{deblur-nerf} are also evaluated. The quantitative results, as presented in Table \ref{tab:nvs_real}, demonstrate the superior performance of our method compared to other approaches. Furthermore, the qualitative results illustrated in Fig. \ref{fig:comparison-real} vividly demonstrate our method's ability to deliver intricate details.

\PAR{Pose Estimation.} Due to the unknown metric scale, we align the estimated trajectories against their corresponding ground truth for the computation of the absolute trajectory error metric. The experimental results presented in Table \ref{table_ate} demonstrate the effectiveness of our approach in recovering precise camera poses.

\section{Conclusion}
We presented the first pipeline to learn Gaussian splattings from a set of motion-blurred images with inaccurate camera poses.
Our pipeline can jointly optimize the 3D scene representation and camera motion trajectories.
Extensive experimental evaluations demonstrate that our method can deliver high-quality novel view images, and achieve real-time rendering compared to prior state-of-the-art works.

\par\vfill\par



\title{BAD-Gaussians: Bundle Adjusted Deblur Gaussian Splatting}
\author{}
\institute{}
\subtitle{Supplementary Materials}
\maketitle
\appendix
\renewcommand\thetable{\Alph{table}}
\renewcommand\thefigure{\Alph{figure}}

\section{Introduction}
In this supplementary material, we present the derivation of the 
analytical jacobian of the Gaussians w.r.t the camera poses and
some additional qualitative and quantitative evaluation of our BAD-Gaussians.

\section{Jacobian of the Gaussians w.r.t the Camera Poses}
In
\begin{equation}
	\frac{\partial \cL}{\partial \bT_i} \coloneq
    \underbrace{
    \sum_{k=0}^{K-1}
        \frac{\partial \cL}{\partial \bB_k} 
	        \cdot \frac{1}{n}
            \sum_{i=0}^{n-1}
                \frac{\partial \bB_k}{\partial \bC_i}}_{\textrm{auto-diff}}
                \frac{\partial \bC_i}{\partial \btheta}
	            \frac{\partial \btheta}{\partial \bT_i}
             \,,
    \label{eq:grad_pose}
\end{equation}
we have
\begin{equation}
    \frac{\partial\btheta}{\partial \bT_i} = 
    \begin{bmatrix}
        \frac{\partial\bmu'}{\partial \bT_i} &
        \bcancel{\frac{\partial\bSigma'}{\partial \bT_i}} &
        \bcancel{\frac{\partial\bc}{\partial \bT_i}} &
        \bcancel{\frac{\partial\bo}{\partial \bT_i}}
    \end{bmatrix}
    \,,
\end{equation}
where color $\bc$ and opacity $\bo$ of the Gaussian are independent 
with the virtual camera pose $\bT_i = \begin{bmatrix}\bR & \bt\end{bmatrix} \in \mathbf{SE}(3)$ of the $i^\textrm{th}$ virtual sharp image $\bC_i$.
Also, following GS-SLAM \cite{yan2023gs}, we ignore 
$\frac{\partial\bSigma'}{\partial \bT_i}$ for efficiency.

As for $\frac{\partial\bmu'}{\partial \bT_i}$, 
since the motion-blurred image synthesis is implemented in PyTorch\cite{paszke2019pytorch}, 
the first part of the gradient in Eq. (\ref{eq:grad_pose}) 
can be computed by the auto-diff module of PyTorch.
The remaining part of the gradient in Eq. (\ref{eq:grad_pose}) can be computed 
as follows:
\begin{equation}
    \frac{\partial \bC_i}{\partial \bmu'}
    \frac{\partial \bmu'}{\partial \bT_i} = 
    \frac{\partial \bC_i}{\partial \bmu'}
    \frac{\partial \bmu'}{\partial \bmu}
    \frac{\partial \bmu}{\partial \bmu_c}
    \frac{\partial \bmu_c}{\partial \bT_i}
    \,,
\end{equation}
where $\bmu_c = \bR \bmu + \bt$ represents $\bmu$ transformed into the camera's coordinate space.

The first term $\frac{\partial \bC_i}{\partial \bmu'}
    \frac{\partial \bmu'}{\partial \bmu}=
    \frac{\partial \bC_i}{\partial \bmu}$ is the Jacobian w.r.t. 
the mean position of the Gaussians, which is already computed in the CUDA backend of the
differentiable projection and rasterization.
The second term can be simplified as follows:
\begin{equation}
    \frac{\partial \bmu}{\partial \bmu_c} = 
    \frac{\partial \bmu}{\partial \bR\bmu + \bt}
     = \bR^\top
\,;
\end{equation}
And the last term:
\begin{equation}
    \frac{\partial \bmu_c}{\partial \bT_i} = 
\mathbf{I}_{3}
\otimes
\begin{bmatrix}
    \bmu ^\top & 1    
\end{bmatrix} \in \mathbb{R}^{3\times12}
\,,
\end{equation}
where $\otimes$ is the Kronecker operator and $\bI_{3}$ is a $3 \times 3$ identity matrix\cite{blanco2010tutorial} \cite{ye2023mathematical}.

Finally, note that the virtual camera pose $\bT_i$ is interpolated from the control knots of the $\mathbf{SE}(3)$ coutinuous trajectory, e.g. $\bT_\textrm{start}, \bT_\textrm{end} \in \mathbf{SE}(3)$ for linear interpolation, and $\bT_1, \bT_2, \bT_3, \bT_4 \in \mathbf{SE}(3)$ for cubic B-spline. We use PyPose\cite{wang2023pypose} to implement the interpolations, thus the corresponding Jacobian of $\bT_i$ w.r.t. the pose adjustments (the actual parameters being optimized), e.g. $\bvarepsilon_\textrm{start}, \bvarepsilon_\textrm{end} \in \mathfrak{se}(3)$ for linear interpolation and $\bvarepsilon_1, \bvarepsilon_2, \bvarepsilon_3, \bvarepsilon_4 \in \mathfrak{se}(3)$ for cubic B-spline, are handled by auto-diff of PyTorch\cite{paszke2019pytorch}.


\section{On Ablation Studies}


\subsection{Full Table of Ablations on Trajectory Representations}
The full results of our ablation study on trajectory representations are presented in Table \ref{tab:nvs_synthetic_full}. The results demonstrate that linear interpolation adequately represents the camera motion trajectory for synthetic datasets, such as \textit{MBA-VO} and \textit{Deblur-NeRF-Synthetic}. However, cubic B-spline outperforms linear interpolation in real data scenarios (\ie \textit{Deblur-NeRF-Real}), attributed to the extended exposure time.

\subsection{Details of Ablations on the Number of Virtual Cameras}
In our ablation study on the number of virtual cameras $n$, for a fair comparison, we make the number of the densified Gaussians roughly the same by adjusting the threshold of the gradient in densification with $n$. This is based on the following derivation: In Eq. (\ref{eq:grad_pose}), during the synthesis of motion-blurred image, the gradient of every Gaussian is scaled by $\frac{1}{n}$. Therefore, if we change $n$ to $n'$, the densification threshold should be multiplied by $\frac{n}{n'}$, in order to match the scaled gradients.

\begin{table*}[!t]
    \centering
    \caption{{\bf{Ablation studies on the effect of trajectory representations.}} The results demonstrate that cubic interpolation improves performance in scenes with complex camera trajectories (\ie {\it{MBA-VO}} and {\it{Deblur-NeRF-Real}}).}
    \setlength\tabcolsep{4pt}{
    \resizebox{0.85\linewidth}{!}{
		\begin{tabular}{c|c|ccc|ccc}
			\hline
			& & \multicolumn{3}{c|}{Linear Interpolation}& \multicolumn{3}{c}{Cubic B-spline} \\
			\textit{Dataset} & Sequence& {\scriptsize PSNR$\uparrow$} & {\scriptsize SSIM$\uparrow$} & {\scriptsize LPIPS$\downarrow$} & {\scriptsize PSNR$\uparrow$} & {\scriptsize SSIM$\uparrow$} & {\scriptsize LPIPS$\downarrow$} \\
			\hline
                \multirow{5}{*}{\begin{tabular}{c}
                    \textit{Deblur-NeRF}\\[\ExtraSep]
                    \textit{Synthetic} \cite{deblur-nerf}
                \end{tabular}}
			&Cozy2room & \cellcolor{red!20}34.68 & \cellcolor{red!20}.9521 & \cellcolor{red!20}.0258 & 33.74 & .9446 & .0346 \\
                &Factory & \cellcolor{red!20}31.88 & .9270 & .0952 & 31.87 & \cellcolor{red!20}.9324 & \cellcolor{red!20}.0897 \\
                &Pool & \cellcolor{red!20}36.95 & \cellcolor{red!20}.9434 & \cellcolor{red!20}.0225 & 34.77 & .9107 & .0440 \\
                &Tanabata & \cellcolor{red!20}32.12 & \cellcolor{red!20}.9481 & \cellcolor{red!20}.0464 & 32.09 & .9477 & \cellcolor{red!20}.0464 \\
                &Trolley & \cellcolor{red!20}33.97 & \cellcolor{red!20}.9628 & \cellcolor{red!20}.0209 & 33.73 & .9619 & .0220 \\
			\hline
                \multirow{2}{*}{\textit{MBA-VO} \cite{mba-vo}}
                &ArchViz-low & 32.28 & \cellcolor{red!20}.9167 & .1134 & \cellcolor{red!20}32.43 & .9165 & \cellcolor{red!20}.1118 \\
                &ArchViz-high & 29.64 & .8568 & .1847 & \cellcolor{red!20}29.68 & \cellcolor{red!20}.8601 & \cellcolor{red!20}.1789 \\
                \hline
                \multirow{10}{*}{\begin{tabular}{c}
                    \textit{Deblur-NeRF}\\[\ExtraSep]
                    \textit{Real} \cite{deblur-nerf}
                \end{tabular}}
                &Ball       & 23.10 & .6423 & .2778 & \cellcolor{red!20}28.11 & \cellcolor{red!20}.8041 & \cellcolor{red!20}.2078 \\
                &Basket     & 27.03 & .8564 & .0998 & \cellcolor{red!20}27.41 & \cellcolor{red!20}.8632 & \cellcolor{red!20}.0987 \\
                &Buick      & 23.44 & .7939 & \cellcolor{red!20}.1118 & \cellcolor{red!20}24.22 & \cellcolor{red!20}.8064 & .1161 \\
                &Coffee     & 30.52 & .9017 & .0913 & \cellcolor{red!20}32.17 & \cellcolor{red!20}.9280 & \cellcolor{red!20}.0659 \\
                &Decoration & 24.99 & .8305 & .0992 & \cellcolor{red!20}25.53 & \cellcolor{red!20}.8442 & \cellcolor{red!20}.0937 \\
                &Girl       & 21.23 & .8028 & .1449 & \cellcolor{red!20}21.28 & \cellcolor{red!20}.8152 & \cellcolor{red!20}.1040 \\
                &Heron      & 21.70 & .6929 & .1495 & \cellcolor{red!20}24.52 & .\cellcolor{red!20}7657 & \cellcolor{red!20}.1187 \\
                &Parterre   & 25.03 & .7817 & .1173 & \cellcolor{red!20}25.94 & \cellcolor{red!20}.8133 & \cellcolor{red!20}.0983 \\
                &Puppet     & 24.78 & .7811 & .1021 & \cellcolor{red!20}25.25 & \cellcolor{red!20}.7991 & \cellcolor{red!20}.0948 \\
                &Stair      & 25.87 & .6944 & .0975 & \cellcolor{red!20}26.63 & \cellcolor{red!20}.7177 & \cellcolor{red!20}.0685 \\
                \hline
            \end{tabular}
	}
}
    \label{tab:nvs_synthetic_full}
\end{table*}

\section{Additional Qualitative Evaluation}
\begin{figure*}[t]
    \setlength\tabcolsep{1pt}
    \centering
    \begin{tabular}{cccccccc}
        &\raisebox{-0.02in}{\rotatebox[origin=t]{90}{\tiny{Input}}} &
        \includegraphics[valign=m,width=0.98\textwidth]{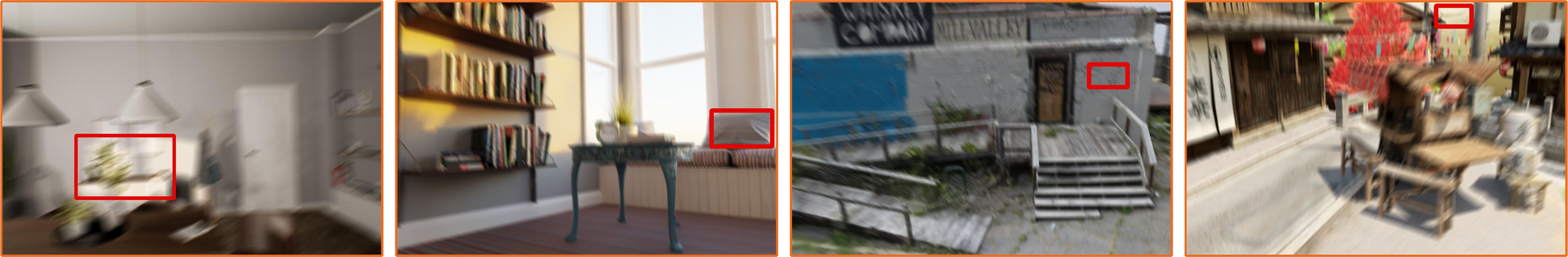}\\
        \specialrule{0em}{.05em}{.05em}
        &\raisebox{-0.02in}{\rotatebox[origin=t]{90}{\tiny{3D-GS\cite{kerbl3Dgaussians}}}} &
        \includegraphics[valign=m,width=0.98\textwidth]{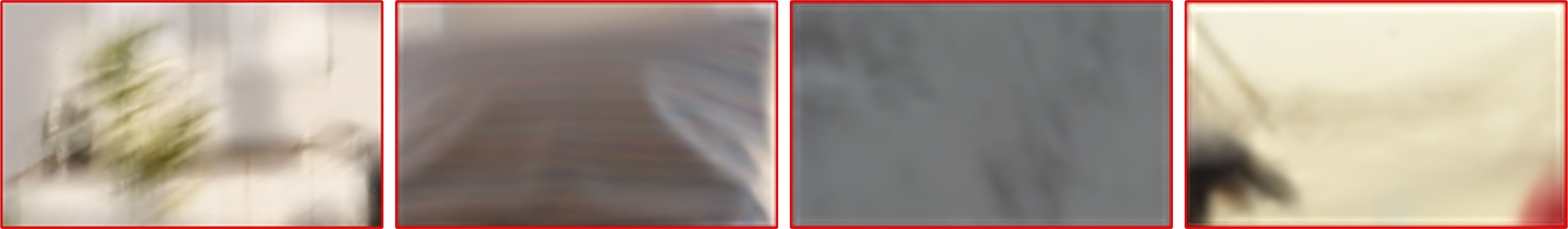}\\
        \specialrule{0em}{.05em}{.05em}
        \raisebox{-0.035in}{\rotatebox[origin=t]{90}{\tiny{Deblur-}}}
        &\raisebox{-0.035in}{\rotatebox[origin=t]{90}{\tiny{NeRF*\cite{deblur-nerf}}}} &
        \includegraphics[valign=m,width=0.98\textwidth]{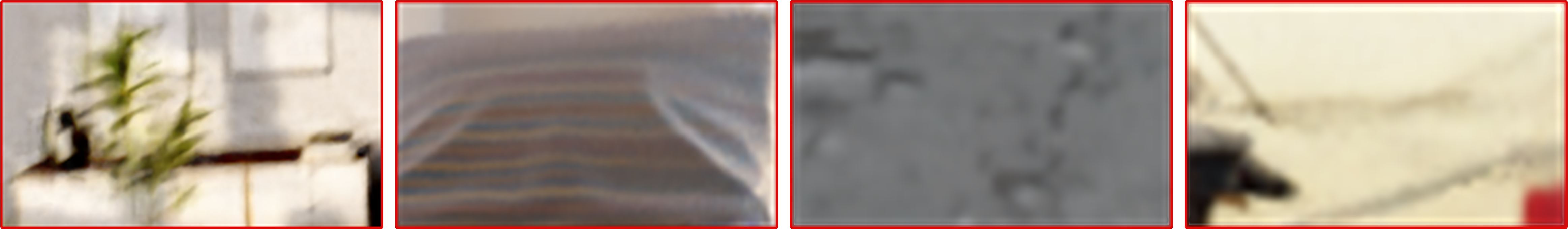}\\
        \specialrule{0em}{.05em}{.05em}
        \raisebox{-0.035in}{\rotatebox[origin=t]{90}{\tiny{DP-}}}
        &\raisebox{-0.035in}{\rotatebox[origin=t]{90}{\tiny{NeRF*\cite{Lee_2023_CVPR}}}} &
        \includegraphics[valign=m,width=0.98\textwidth]{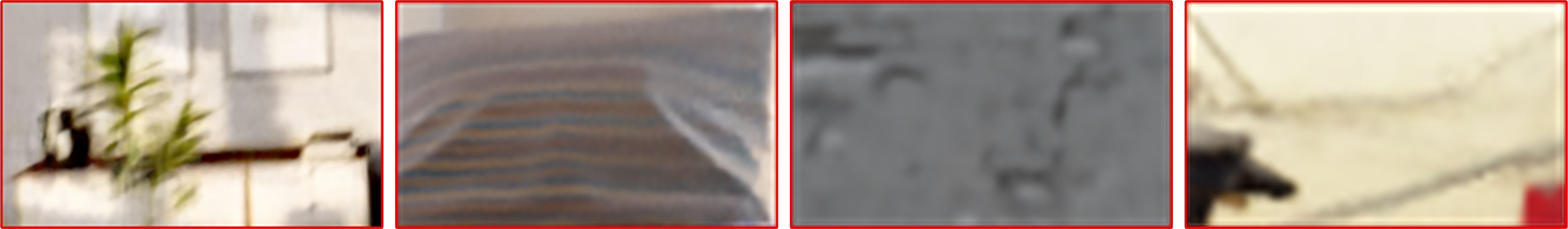}\\
        \specialrule{0em}{.05em}{.05em}
        \raisebox{-0.035in}{\rotatebox[origin=t]{90}{\tiny{BAD-}}}
        &\raisebox{-0.035in}{\rotatebox[origin=t]{90}{\tiny{NeRF\cite{wang2023badnerf}}}} &
        \includegraphics[valign=m,width=0.98\textwidth]{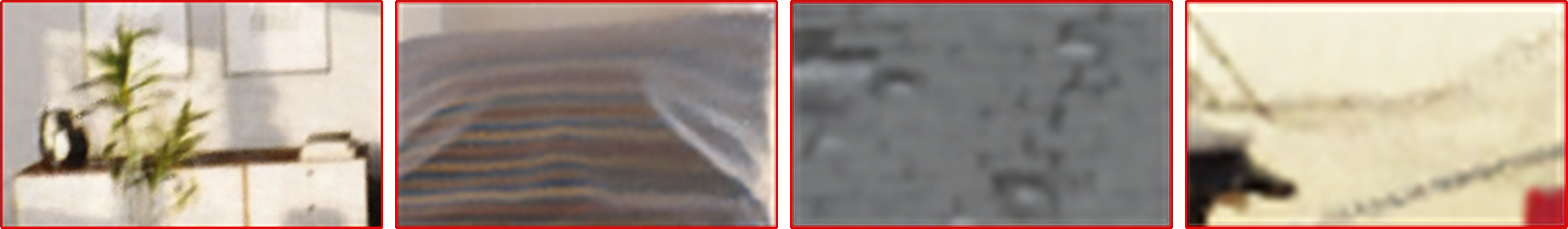}\\
        \specialrule{0em}{.05em}{.05em}
        &\raisebox{-0.035in}{\rotatebox[origin=t]{90}{\scriptsize Ours}} &
        \includegraphics[valign=m,width=0.98\textwidth]{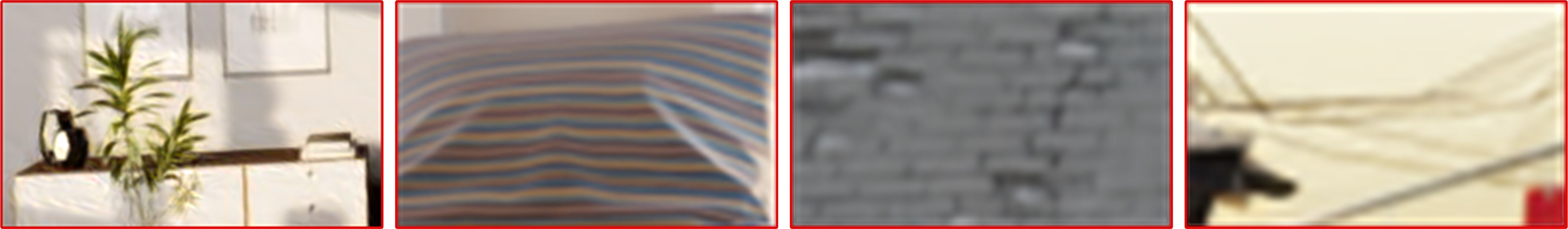}\\
        \specialrule{0em}{.05em}{.05em}
        &\raisebox{-0.035in}{\rotatebox[origin=t]{90}{\tiny{Reference}}} &
        \includegraphics[valign=m,width=0.98\textwidth]{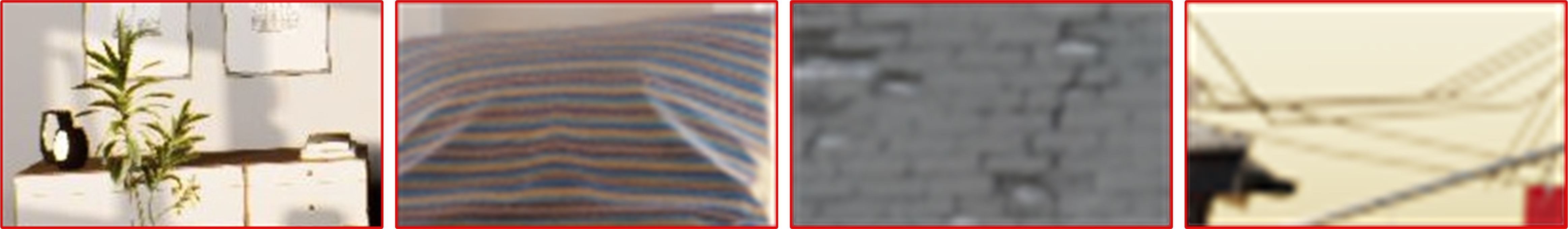}\\
    \end{tabular}
    \caption{ {\textbf{Qualitative deblurring results of different methods with synthetic datasets from MBA-VO \cite{mba-vo} and Deblur-NeRF \cite{deblur-nerf}.} The scenes, from left to right, encompass \textit{ArchViz-high}, \textit{Cozy2room}, \textit{Factory}, and \textit{Trolley}. Despite being trained with ground truth poses (*), BAD-Gaussians outperforms Deblur-NeRF* and DP-NeRF* in recovering high-quality scenes from motion-blurred images with inaccurate camera poses, showcasing its superior performance.}}
    \label{fig:comparison-syn-sup}
\end{figure*}

\begin{figure*}[t]
    \setlength\tabcolsep{1pt}
    \centering
    \begin{tabular}{cccccccc}
        &\raisebox{-0.02in}{\rotatebox[origin=t]{90}{\tiny{Novel View}}} &
        \includegraphics[valign=m,width=0.98\textwidth]{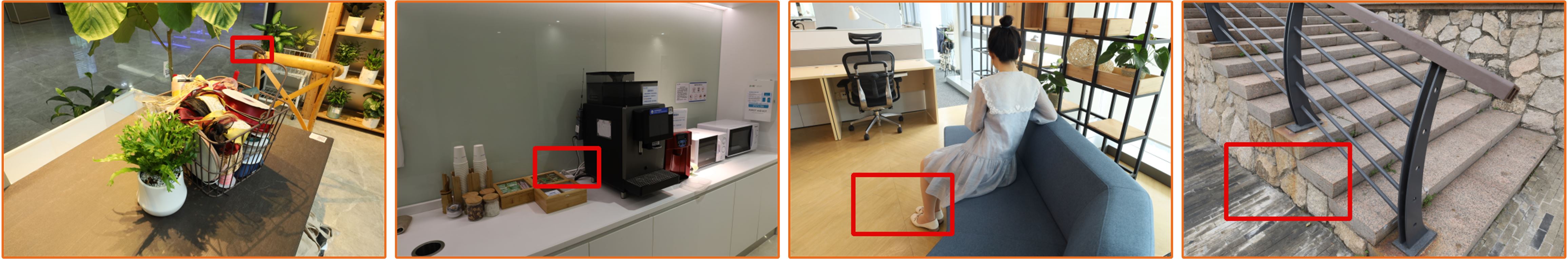}\\
        \specialrule{0em}{.05em}{.05em}
        &\raisebox{-0.02in}{\rotatebox[origin=t]{90}{\tiny{3D-GS\cite{kerbl3Dgaussians}}}} &
        \includegraphics[valign=m,width=0.98\textwidth]{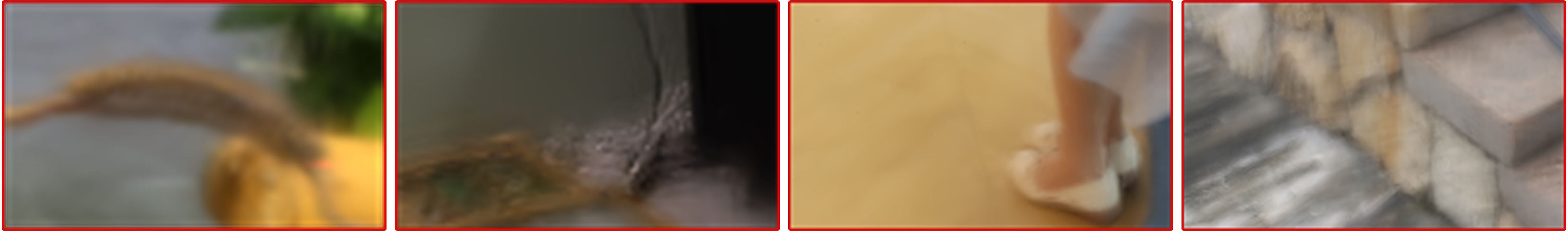}\\
        \specialrule{0em}{.05em}{.05em}
        \raisebox{-0.035in}{\rotatebox[origin=t]{90}{\tiny{Deblur-}}}
        &\raisebox{-0.035in}{\rotatebox[origin=t]{90}{\tiny{NeRF*\cite{deblur-nerf}}}} &
        \includegraphics[valign=m,width=0.98\textwidth]{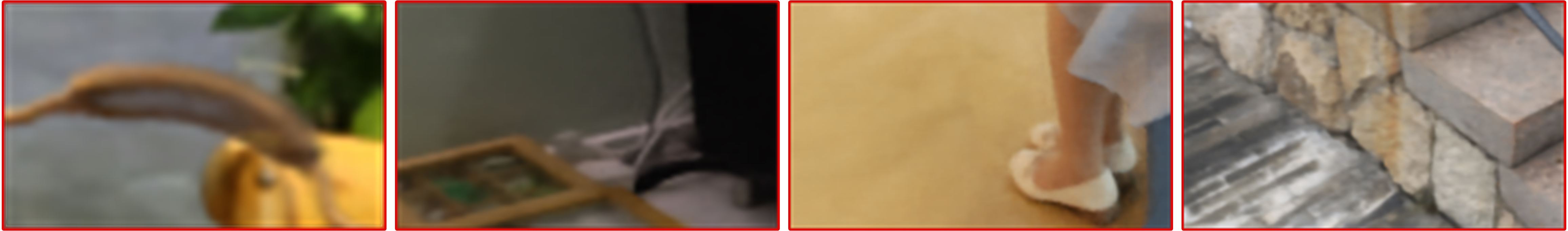}\\
        \specialrule{0em}{.05em}{.05em}
        \raisebox{-0.035in}{\rotatebox[origin=t]{90}{\tiny{DP-}}}
        &\raisebox{-0.035in}{\rotatebox[origin=t]{90}{\tiny{NeRF*\cite{Lee_2023_CVPR}}}} &
        \includegraphics[valign=m,width=0.98\textwidth]{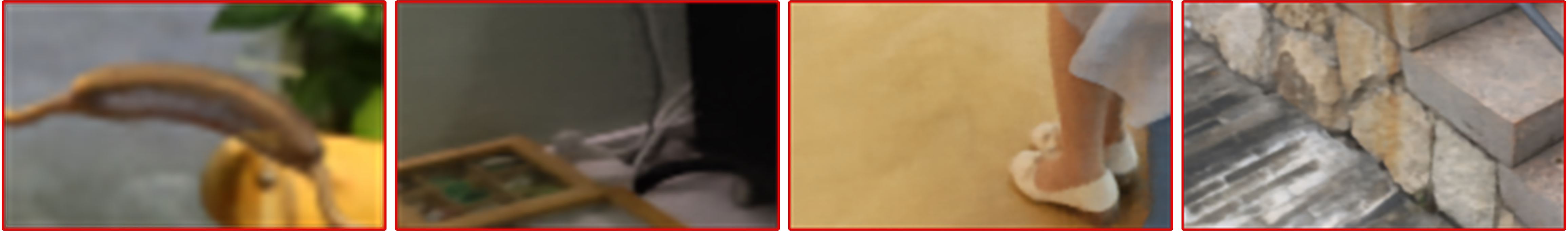}\\
        \specialrule{0em}{.05em}{.05em}
        \raisebox{-0.035in}{\rotatebox[origin=t]{90}{\tiny{BAD-}}}
        &\raisebox{-0.035in}{\rotatebox[origin=t]{90}{\tiny{NeRF\cite{wang2023badnerf}}}} &
        \includegraphics[valign=m,width=0.98\textwidth]{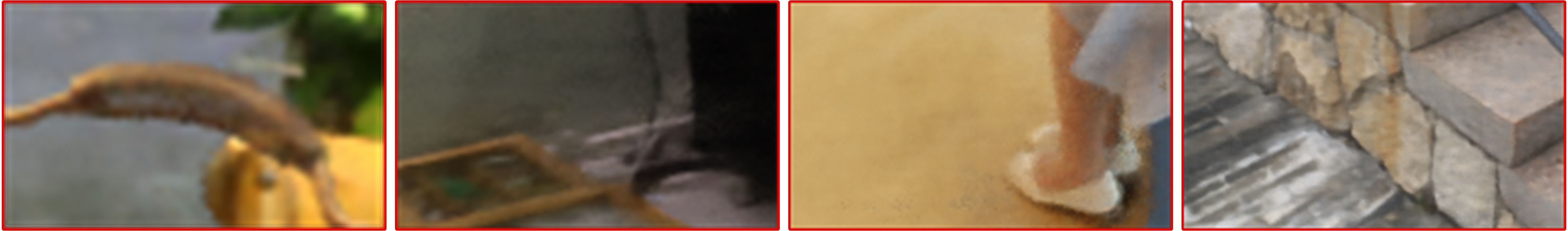}\\
        \specialrule{0em}{.05em}{.05em}
        &\raisebox{-0.035in}{\rotatebox[origin=t]{90}{\scriptsize Ours}} &
        \includegraphics[valign=m,width=0.98\textwidth]{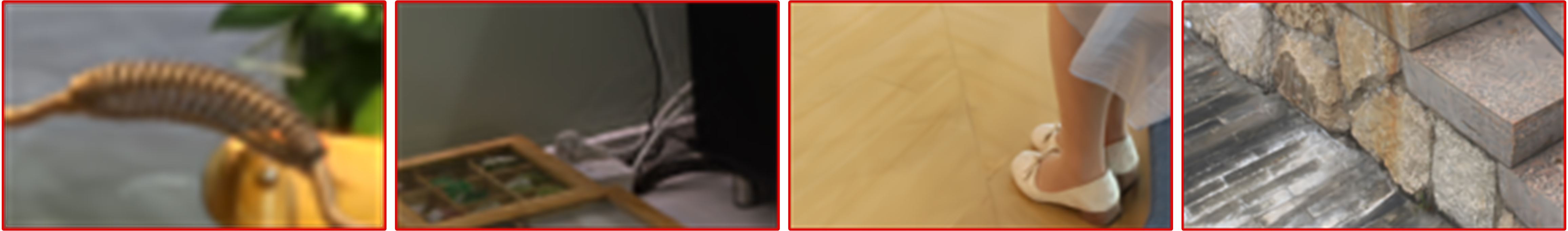}\\
        \specialrule{0em}{.05em}{.05em}
        &\raisebox{-0.035in}{\rotatebox[origin=t]{90}{\tiny{Reference}}} &
        \includegraphics[valign=m,width=0.98\textwidth]{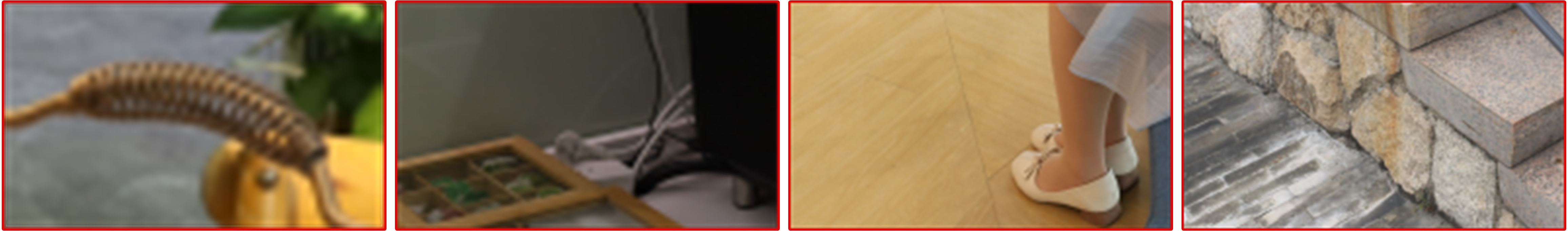}\\
    \end{tabular}
    \caption{{\textbf{Qualitative novel view synthesis results of different methods with the real datasets from Deblur-NeRF \cite{deblur-nerf}.}} The scenes, from left to right, encompass \textit{Basket}, \textit{Coffee}, \textit{Girl}, and \textit{Stair}. The experimental results demonstrate that our method achieves superior performance over prior methods on the real dataset as well.}
    \label{fig:comparison-real-sup}
\end{figure*}

We provide further qualitative experimental results on both the synthetic and real datasets, showcased in Fig. \ref{fig:comparison-syn-sup} and Fig. \ref{fig:comparison-real-sup} respectively. These results demonstrate the superior performance of our method over previous state-of-the-art approaches.

\subsection{Visualization of Pose Estimation Results}

\begin{figure*}[h]
	\setlength\tabcolsep{1pt}
	\centering
	\begin{tabular}{ccc}
		\includegraphics[width=0.333\textwidth]{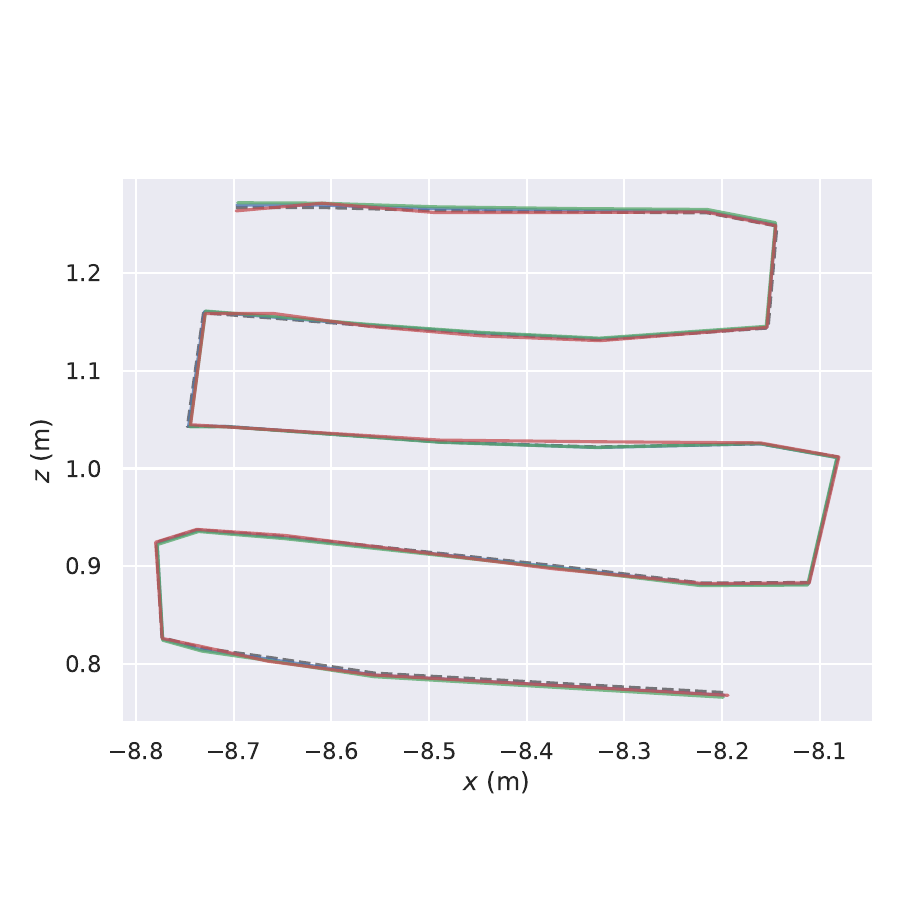} &
		\includegraphics[width=0.333\textwidth]{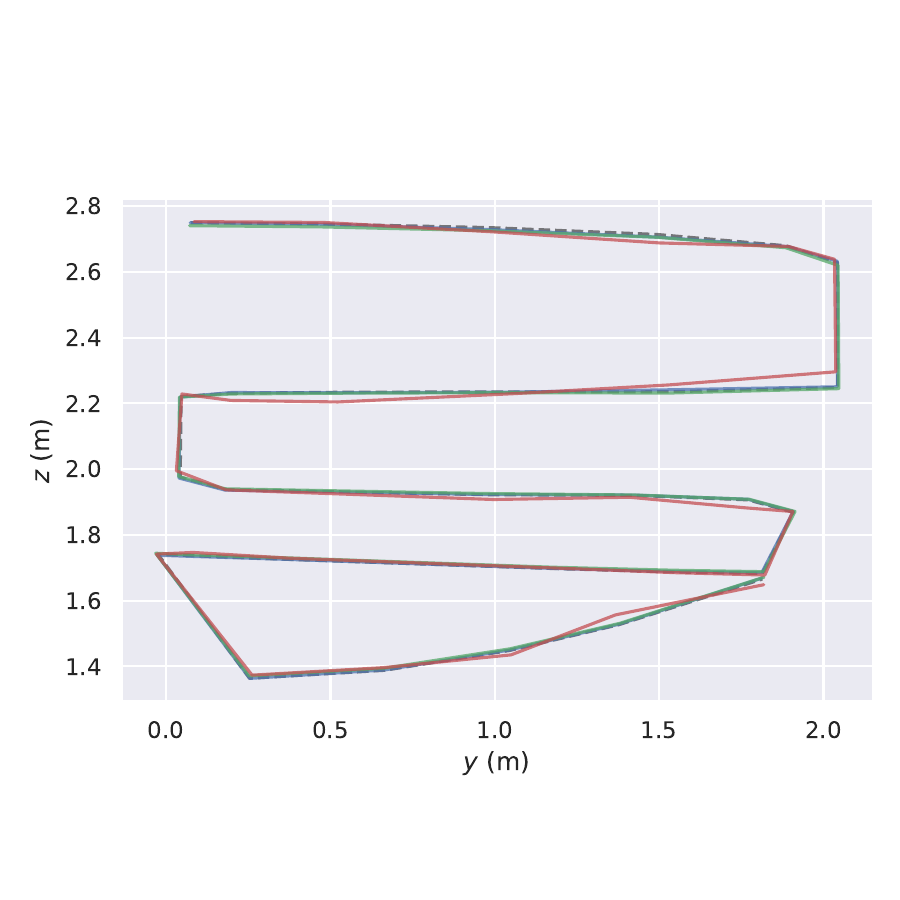} &
		\includegraphics[width=0.333\textwidth]{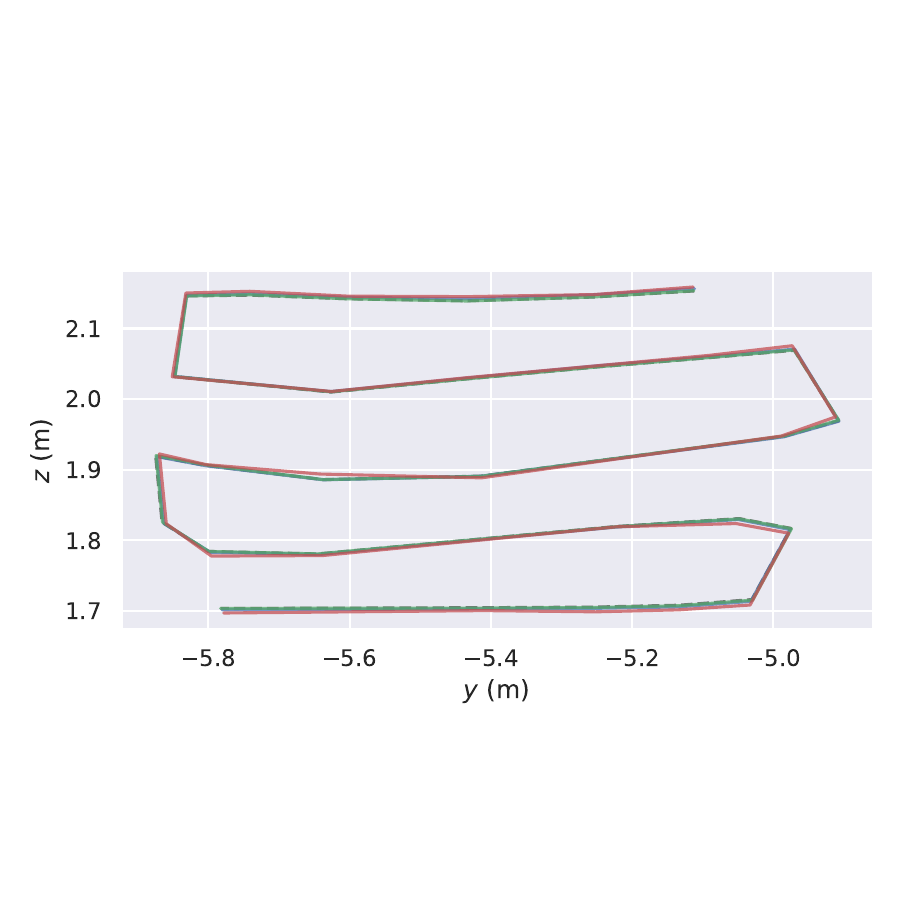} \\
		\includegraphics[width=0.333\textwidth]{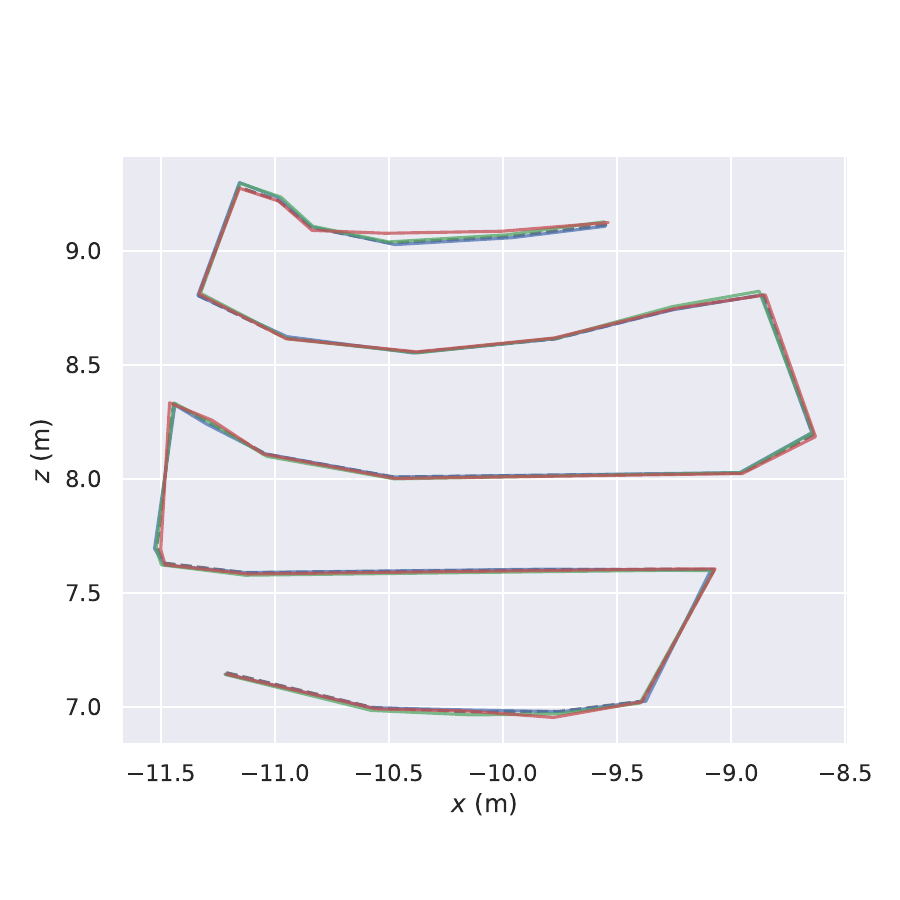} &
		\includegraphics[width=0.333\textwidth]{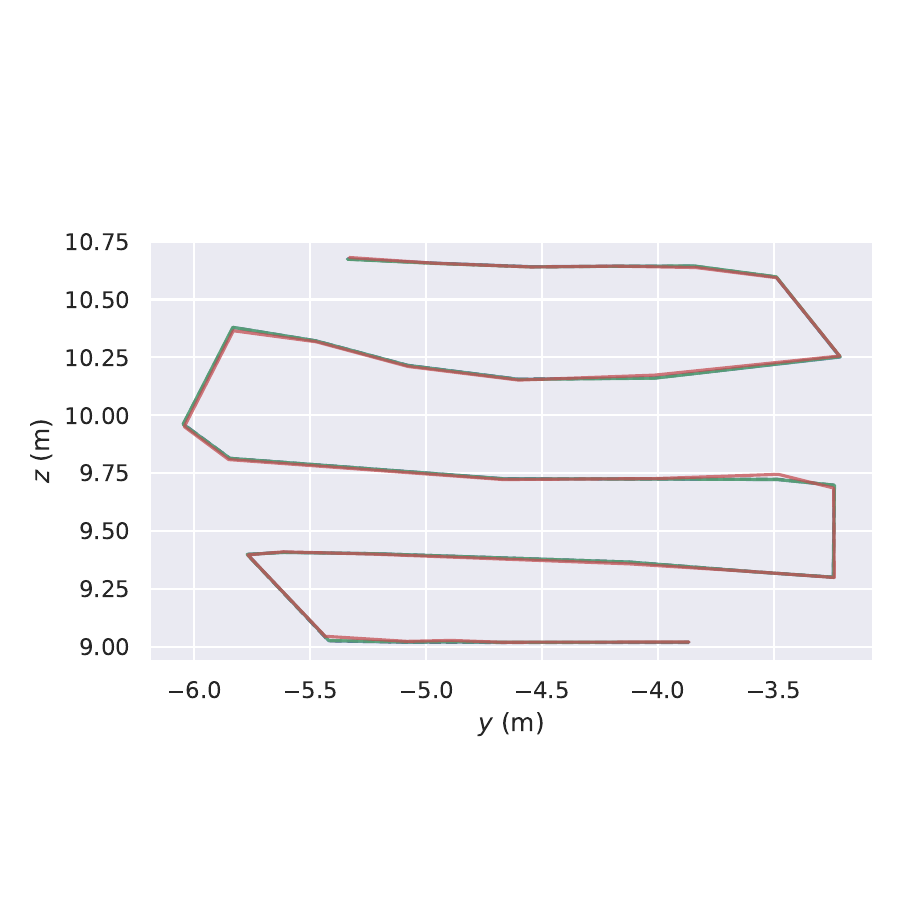} &
		\includegraphics[width=0.333\textwidth]{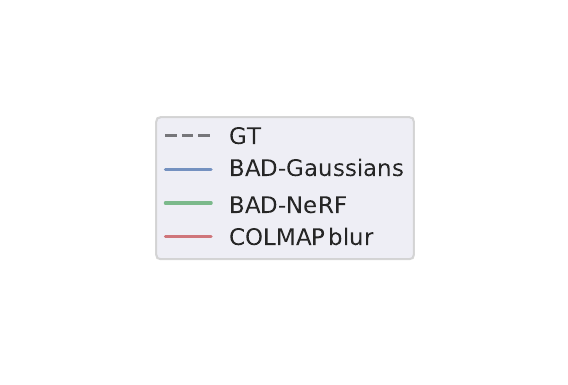} \\
		%
	\end{tabular}
	\vspace{-0.7em}
	\caption{{\textbf{Qualitative Comparisons of estimated camera poses on Deblur-NeRF dataset.}} These are results on \textit{Cozy2room}, \textit{Factory}, \textit{Pool}, \textit{Tanabata} and \textit{Trolley} sequences respectively. The results demonstrate that our method recovers motion trajectories more accurately compared with both COLMAP \cite{colmap} and BAD-NeRF \cite{wang2023badnerf}.}
    \label{fig:trajectory}
\end{figure*}

In this section, we present the visualization results in terms of camera pose estimation. The experiments are conducted on the synthetic dataset of Deblur-NeRF \cite{deblur-nerf}. We present the comparison result of BAD-Gaussians against COLMAP \cite{colmap} and BAD-NeRF \cite{wang2023badnerf} in \figref{fig:trajectory}. It demonstrates that our method recovers motion trajectories more accurately.

\subsection{Video of Novel View Synthesis Results}
To showcase the effectiveness of our approach, we provide supplementary videos illustrating the capability of BAD-Gaussians to recover high-quality latent sharp video from blurry images. The videos contain results on both synthetic and real scenes from Deblur-NeRF\cite{deblur-nerf}. In the provided videos, on the left are our rendered novel view images and on the right are the input blurry images.

Notably, in the provided videos, due to the fast training speed and low GPU memory requirements of our method, we are able to train real scenes at the native resolution $2400\times 1600$ to achieve maximum reconstruction quality in about 1.5 hours, compared to the resolution of $600\times 400 $ that we used in all experiments above in this paper for a fair comparison.

\clearpage
%

%
\bibliographystyle{splncs04}
\bibliography{reference}

\begin{thebibliography}{10}
\providecommand{\url}[1]{\texttt{#1}}
\providecommand{\urlprefix}{URL }
\providecommand{\doi}[1]{https://doi.org/#1}

\bibitem{mipnerf}
Barron, J.T., Mildenhall, B., Tancik, M., Hedman, P., Martin-Brualla, R., Srinivasan, P.P.: {Mip-NeRF: A Multiscale Representation for Anti-Aliasing Neural Radiance Fields}. In: International Conference on Computer Vision (ICCV) (2021), \url{https://jonbarron.info/mipnerf/}

\bibitem{barron2022mipnerf360}
Barron, J.T., Mildenhall, B., Verbin, D., Srinivasan, P.P., Hedman, P.: {Mip-NeRF 360: Unbounded Anti-Aliased Neural Radiance Fields}. In: Computer Vision and Pattern Recognition (CVPR) (2022), \url{https://jonbarron.info/mipnerf360/}

\bibitem{barron2023zipnerf}
Barron, J.T., Mildenhall, B., Verbin, D., Srinivasan, P.P., Hedman, P.: {Zip-NeRF: Anti-Aliased Grid-Based Neural Radiance Fields}. In: International Conference on Computer Vision (ICCV) (2023), \url{https://jonbarron.info/zipnerf/}

\bibitem{blanco2010tutorial}
Blanco, J.L.: {A tutorial on SE(3) transformation parameterizations and on-manifold optimization}. Tech. Rep. 012010, University of Malaga (2010), \url{http://ingmec.ual.es/~jlblanco/papers/jlblanco2010geometry3D_techrep.pdf}

\bibitem{Cao2023HEXPLANE}
Cao, A., Johnson, J.: {HexPlane: A Fast Representation for Dynamic Scenes}. In: Computer Vision and Pattern Recognition (CVPR) (2023), \url{https://caoang327.github.io/HexPlane/}

\bibitem{Chen2022ECCV}
Chen, A., Xu, Z., Geiger, A., Yu, J., Su, H.: {TensoRF: Tensorial Radiance Fields}. In: European Conference on Computer Vision (ECCV) (2022), \url{https://apchenstu.github.io/TensoRF/}

\bibitem{Cho2009ToG}
Cho, S., Lee, S.: {Fast Motion Deblurring}. ACM Transactions on Graphics (TOG)  \textbf{28}(5) (2009)

\bibitem{Blender}
Community, B.O.: {Blender - a 3D modelling and rendering package}. Blender Foundation, Stichting Blender Foundation, Amsterdam (2018), \url{http://www.blender.org}

\bibitem{depthnerf}
Deng, K., Liu, A., Zhu, J.Y., Ramanan, D.: {Depth-supervised NeRF: Fewer Views and Faster Training for Free}. In: Computer Vision and Pattern Recognition (CVPR) (2022), \url{https://www.cs.cmu.edu/~dsnerf/}

\bibitem{Fergus2006SIGGGRAPH}
Fergus, R., Singh, B., Hertzmann, A., Roweis, S.T., Freeman, W.T.: {Removing camera shake from a single photograph}. In: ACM Transactions on Graphics (TOG) (2006)

\bibitem{kplanes_2023}
Fridovich-Keil, S., Meanti, G., Warburg, F.R., Recht, B., Kanazawa, A.: {K-Planes: Explicit Radiance Fields in Space, Time, and Appearance}. In: Computer Vision and Pattern Recognition (CVPR) (2023), \url{https://sarafridov.github.io/K-Planes/}

\bibitem{fu2023colmap}
Fu, Y., Liu, S., Kulkarni, A., Kautz, J., Efros, A.A., Wang, X.: {Colmap-Free 3D Gaussian Splatting}. arXiv preprint arXiv:2312.07504  (2023), \url{https://oasisyang.github.io/colmap-free-3dgs/}

\bibitem{Garbin2021}
Garbin, S.J., Kowalski, M., Johnson, M., Shotton, J., Valentin, J.: {FastNeRF: High-Fidelity Neural Rendering at 200FPS}. In: International Conference on Computer Vision (ICCV) (2021), \url{https://microsoft.github.io/FastNeRF/}

\bibitem{Jeong2021}
Jeong, Y., Ahn, S., Choy, C., Anandkumar, A., Cho, M., Park, J.: {Self-Calibrating Neural Radiance Fields}. In: International Conference on Computer Vision (ICCV) (2021), \url{https://postech-cvlab.github.io/SCNeRF/}

\bibitem{kerbl3Dgaussians}
Kerbl, B., Kopanas, G., Leimk{\"u}hler, T., Drettakis, G.: {3D Gaussian Splatting for Real-Time Radiance Field Rendering}. ACM Transactions on Graphics (TOG)  \textbf{42}(4) (July 2023), \url{https://repo-sam.inria.fr/fungraph/3d-gaussian-splatting/}

\bibitem{infonerf}
Kim, M., Seo, S., Han, B.: {InfoNeRF: Ray Entropy Minimization for Few-Shot Neural Volume Rendering}. In: Computer Vision and Pattern Recognition (CVPR) (2022), \url{https://cv.snu.ac.kr/research/InfoNeRF/}

\bibitem{Krishnan2009NIPS}
Krishnan, D., Fergus, R.: {Fast image deconvolution using Hyper-Laplacian priors}. In: Neural Information Processing Systems (NIPS) (2009)

\bibitem{Kupyn2019ICCV}
Kupyn, O., Martyniuk, T., Wu, J., Wang, Z.: {DeblurGAN-v2: Deblurring (Orders-of-Magnitude) Faster and Better}. In: International Conference on Computer Vision (ICCV) (2019), \url{https://github.com/VITA-Group/DeblurGANv2}

\bibitem{Lassner2021}
Lassner, C., Zollhofer, M.: {Pulsar: Efficient sphere-based neural rendering}. In: Computer Vision and Pattern Recognition (CVPR) (2021), \url{https://github.com/facebookresearch/pytorch3d/}

\bibitem{Lee_2023_CVPR}
Lee, D., Lee, M., Shin, C., Lee, S.: {DP-NeRF: Deblurred Neural Radiance Field With Physical Scene Priors}. In: Computer Vision and Pattern Recognition (CVPR) (2023), \url{https://dogyoonlee.github.io/dpnerf/}

\bibitem{Levin2009CVPR}
Levin, A., Weiss, Y., Durand, F., Freeman, W.T.: {Understanding and evaluating blind deconvolution algorithm}. In: Computer Vision and Pattern Recognition (CVPR) (2009)

\bibitem{levoy1990efficient}
Levoy, M.: {Efficient Ray Tracing of Volume Data}. ACM Transactions on Graphics (TOG)  \textbf{9}(3),  245--261 (1990)

\bibitem{li2023usb}
Li, M., Wang, P., Zhao, L., Liao, B., Liu, P.: {USB-NeRF: Unrolling Shutter Bundle Adjusted Neural Radiance Fields}. In: International Conference on Learning Representations (ICLR) (2024), \url{https://moyangli00.github.io/usb_nerf/}

\bibitem{Lin2021}
{Lin, Chen-Hsuan and Ma, Wei-Chiu and Torralba, Antonio and Lucey, Simon}: {BARF: Bundle-Adjusting Neural Radiance Fields}. In: International Conference on Computer Vision (ICCV) (2021), \url{https://chenhsuanlin.bitbucket.io/bundle-adjusting-NeRF}

\bibitem{mba-vo}
Liu, P., Zuo, X., Larsson, V., Pollefeys, M.: {MBA-VO: Motion Blur Aware Visual Odometry}. In: International Conference on Computer Vision (ICCV) (2021), \url{https://github.com/ethliup/MBA-VO}

\bibitem{deblur-nerf}
Ma, L., Li, X., Liao, J., Zhang, Q., Wang, X., Wang, J., Sander, P.V.: {Deblur-NeRF: Neural Radiance Fields from Blurry Images}. In: Computer Vision and Pattern Recognition (CVPR) (2022), \url{https://limacv.github.io/deblurnerf/}

\bibitem{max1995optical}
Max, N.: {Optical Models for Direct Volume Rendering}. ACM Transactions on Graphics (TOG)  \textbf{1}(2),  99--108 (1995)

\bibitem{mildenhall2022rawnerf}
Mildenhall, B., Hedman, P., Martin-Brualla, R., Srinivasan, P.P., Barron, J.T.: {NeRF} in the dark: High dynamic range view synthesis from noisy raw images. In: Computer Vision and Pattern Recognition (CVPR) (2022), \url{https://bmild.github.io/rawnerf/}

\bibitem{nerf}
Mildenhall, B., Srinivasan, P.P., Tancik, M., Barron, J.T., Ramamoorthi, R., Ng, R.: {NeRF: Representing Scenes as Neural Radiance Fields for View Synthesis}. In: European Conference on Computer Vision (ECCV) (2020), \url{https://www.matthewtancik.com/nerf}

\bibitem{mueller2022instant}
M\"uller, T., Evans, A., Schied, C., Keller, A.: {Instant Neural Graphics Primitives with a Multiresolution Hash Encoding}. ACM Transactions on Graphics (TOG)  \textbf{41}(4),  102:1--102:15 (Jul 2022), \url{https://nvlabs.github.io/instant-ngp/}

\bibitem{Nah2017CVPR}
Nah, S., Kim, T.H., Lee, K.M.: {Deep Multi-Scale Convolutional Neural Network for Dynamic Scene Deblurring}. In: Computer Vision and Pattern Recognition (CVPR) (2017), \url{https://github.com/SeungjunNah/DeepDeblur_release}

\bibitem{regnerf}
Niemeyer, M., Barron, J.T., Mildenhall, B., Sajjadi, M.S., Geiger, A., Radwan, N.: {RegNeRF: Regularizing Neural Radiance Fields for View Synthesis from Sparse Inputs}. In: Computer Vision and Pattern Recognition (CVPR) (2022), \url{https://m-niemeyer.github.io/regnerf/}

\bibitem{Park2017}
Park, H., Lee, K.M.: {Joint Estimation of Camera Pose, Depth, Deblurring, and Super-Resolution from a Blurred Image Sequence}. In: International Conference on Computer Vision (ICCV). vol. 2017-Octob (2017). \doi{10.1109/ICCV.2017.494}

\bibitem{park2023camp}
Park, K., Henzler, P., Mildenhall, B., Barron, J.T., Martin-Brualla, R.: {CamP: Camera Preconditioning for Neural Radiance Fields}. ACM Transactions on Graphics (TOG)  (2023), \url{https://camp-nerf.github.io/}

\bibitem{paszke2019pytorch}
Paszke, A., Gross, S., Massa, F., Lerer, A., Bradbury, J., Chanan, G., Killeen, T., Lin, Z., Gimelshein, N., Antiga, L., et~al.: {Pytorch: An Imperative Style, High-performance Deep Learning Library}. Advances in neural information processing systems  \textbf{32} (2019), \url{https://pytorch.org/}

\bibitem{Piala2022}
Piala, M., Clark, R.: {TermiNeRF: Ray Termination Prediction for Efficient Neural Rendering}. In: International Conference on 3D Vision (3DV) (2021), \url{https://projects.mackopes.com/terminerf/}

\bibitem{yu_and_fridovichkeil2021plenoxels}
{Sara Fridovich-Keil and Alex Yu}, Tancik, M., Chen, Q., Recht, B., Kanazawa, A.: {Plenoxels: Radiance Fields without Neural Networks}. In: Computer Vision and Pattern Recognition (CVPR) (2022), \url{https://alexyu.net/plenoxels/}

\bibitem{colmap}
Schonberger, J.L., Frahm, J.M.: {Structure-from-motion Revisited}. In: Computer Vision and Pattern Recognition (CVPR) (2016), \url{https://github.com/colmap/colmap}

\bibitem{schops2019bad}
Schops, T., Sattler, T., Pollefeys, M.: {BAD SLAM: Bundle Adjusted Direct RGB-D SLAM}. In: Computer Vision and Pattern Recognition (CVPR) (2019), \url{https://github.com/ETH3D/badslam}

\bibitem{Shan2008ToG}
Shan, Q., Jia, J., A., A.: {High-quality Motion Deblurring from a Single Image}. ACM Transactions on Graphics (TOG)  \textbf{27}(3) (2008)

\bibitem{Su2017CVPR}
Su, S., Delbracio, M., Wang, J.: {Deep Video Deblurring for Hand-held Cameras}. In: Computer Vision and Pattern Recognition (CVPR) (2017), \url{https://github.com/shuochsu/DeepVideoDeblurring}

\bibitem{SunSC22}
Sun, C., Sun, M., Chen, H.: {Direct Voxel Grid Optimization: Super-fast Convergence for Radiance Fields Reconstruction}. In: Computer Vision and Pattern Recognition (CVPR) (2022), \url{https://sunset1995.github.io/dvgo/}

\bibitem{Tao2018CVPR}
Tao, X., Gao, H., Shen, X., Wang, J., Jia, J.: {Scale-recurrent network for deep image deblurring}. In: Computer Vision and Pattern Recognition (CVPR) (2018), \url{https://github.com/jiangsutx/SRN-Deblur}

\bibitem{wang2023pypose}
Wang, C., Gao, D., Xu, K., Geng, J., Hu, Y., Qiu, Y., Li, B., Yang, F., Moon, B., Pandey, A., Aryan, Xu, J., Wu, T., He, H., Huang, D., Ren, Z., Zhao, S., Fu, T., Reddy, P., Lin, X., Wang, W., Shi, J., Talak, R., Cao, K., Du, Y., Wang, H., Yu, H., Wang, S., Chen, S., Kashyap, A., Bandaru, R., Dantu, K., Wu, J., Xie, L., Carlone, L., Hutter, M., Scherer, S.: {PyPose}: A library for robot learning with physics-based optimization. In: IEEE/CVF Conference on Computer Vision and Pattern Recognition (CVPR) (2023), \url{https://github.com/pypose/pypose}

\bibitem{wangsupplementary}
Wang, P., Zhao, L., Ma, R., Liu, P.: {Supplementary Material for BAD-NeRF: Bundle Adjusted Deblur Neural Radiance Fields} \url{https://openaccess.thecvf.com/content/CVPR2023/supplemental/Wang_BAD-NeRF_Bundle_Adjusted_CVPR_2023_supplemental.pdf}

\bibitem{wang2023badnerf}
Wang, P., Zhao, L., Ma, R., Liu, P.: {BAD-NeRF: Bundle Adjusted Deblur Neural Radiance Fields}. In: Computer Vision and Pattern Recognition (CVPR) (2023), \url{https://wangpeng000.github.io/BAD-NeRF/}

\bibitem{wang2021nerfmm}
Wang, Z., Wu, S., Xie, W., Chen, M., Prisacariu, V.A.: Ne{RF}$--$: Neural radiance fields without known camera parameters. arXiv preprint arXiv:2102.07064  (2021), \url{https://nerfmm.active.vision/}

\bibitem{Wizadwongsa2021}
Wizadwongsa, S., Phongthawee, P., Yenphraphai, J., Suwajanakorn, S.: {NeX: Real-time View Synthesis with Neural Basis Expansion}. In: Computer Vision and Pattern Recognition (CVPR) (2021), \url{https://nex-mpi.github.io/}

\bibitem{Xu2010ECCV}
Xu, L., Jia, J.: {Two-Phase Kernel Estimation for robust motion deblurring}. In: European Conference on Computer Vision (ECCV) (2010)

\bibitem{yan2023gs}
Yan, C., Qu, D., Wang, D., Xu, D., Wang, Z., Zhao, B., Li, X.: {GS-SLAM: Dense Visual SLAM with 3D Gaussian Splatting}. arXiv preprint arXiv:2311.11700  (2023)

\bibitem{ye2023mathematical}
Ye, V., Kanazawa, A.: Mathematical supplement for the gsplat library. arXiv preprint arXiv:2312.02121  (2023), \url{https://github.com/nerfstudio-project/gsplat}

\bibitem{Yu2021}
Yu, A., Li, R., Tancik, M., Li, H., Ng, R., Kanazawa, A.: {PlenOctrees for Real-time Rendering of Neural Radiance Fields}. In: International Conference on Computer Vision (ICCV) (2021), \url{https://alexyu.net/plenoctrees/}

\bibitem{zamir2021multi}
Zamir, S.W., Arora, A., Khan, S., Hayat, M., Khan, F.S., Yang, M.H., Shao, L.: {Multi-stage Progressive Image Restoration}. In: Computer Vision and Pattern Recognition (CVPR) (2021), \url{https://github.com/swz30/MPRNet}

\bibitem{zwicker2001ewa}
Zwicker, M., Pfister, H., Van~Baar, J., Gross, M.: {EWA Volume Splatting}. In: Proceedings Visualization, 2001. VIS'01. pp. 29--538. IEEE (2001), \url{https://cgl.ethz.ch/research/past_projects/surfels/ewavolumesplatting/index.html}

\end{thebibliography}
\end{document}